\documentclass[11pt]{article}

\usepackage[final]{latex/acl}

\usepackage{times}
\usepackage{latexsym}

\usepackage[T1]{fontenc}

\usepackage[utf8]{inputenc}

\usepackage{microtype}

\usepackage{inconsolata}

\usepackage{graphicx}

\usepackage{amsmath}
\usepackage{amsfonts}
\usepackage{booktabs}
\usepackage{graphicx}
\usepackage{subcaption}
\usepackage{tabularx}
\usepackage{xcolor}
\usepackage{lscape}

\title{Sieve and Sage: \\ Efficient Distraction Filtering for Reliable RALM Abstention}

\author{Jongbin Won \and Sung Geun An \and Jay-Yoon Lee\thanks{Corresponding author.} \\
  Graduate School of Data Science / Seoul National University \\
  \texttt{\{jongbinwon, ssunggun2, lee.jayyoon\}@snu.ac.kr}
}

\begin{document}
\maketitle
\begin{abstract}
Just as Socrates recognized the limits of his own knowledge, Retrieval-Augmented Language Models (RALMs) should learn to \emph{abstain} when the retrieved evidence cannot support a reliable response. Existing approaches largely rely on monolithic LLMs to handle heterogeneous retrieval failures in a single step, resulting in limited abstention performance and high computational costs. We instead decompose retrieval failures into two distinct states: (i) the \emph{unanswerable state}, where the required evidence is absent, and (ii) the \emph{distracted state}, where relevant evidence is mixed with conflicting, negated, or adversarial information. Based on this decomposition, we introduce a lightweight module (\textbf{Sieve}) that screens retrieved document sets for distracting evidence before invoking a costly LLM (\textbf{Sage}) for grounded generation and abstention. Evaluated across both general and high-stakes expert domains, our Sieve and Sage framework preemptively detects distracting noise, improving system accuracy by up to 69.4\%p and Macro-F1 by 55.2\%p compared to one-stage baselines. Furthermore, it achieves up to a 1.99$\times$ speedup, establishing a highly efficient and reliable abstention pipeline for RALM with abstention.
\end{abstract}

\begin{figure*}
    \centering
    \includegraphics[width=1.0\linewidth]{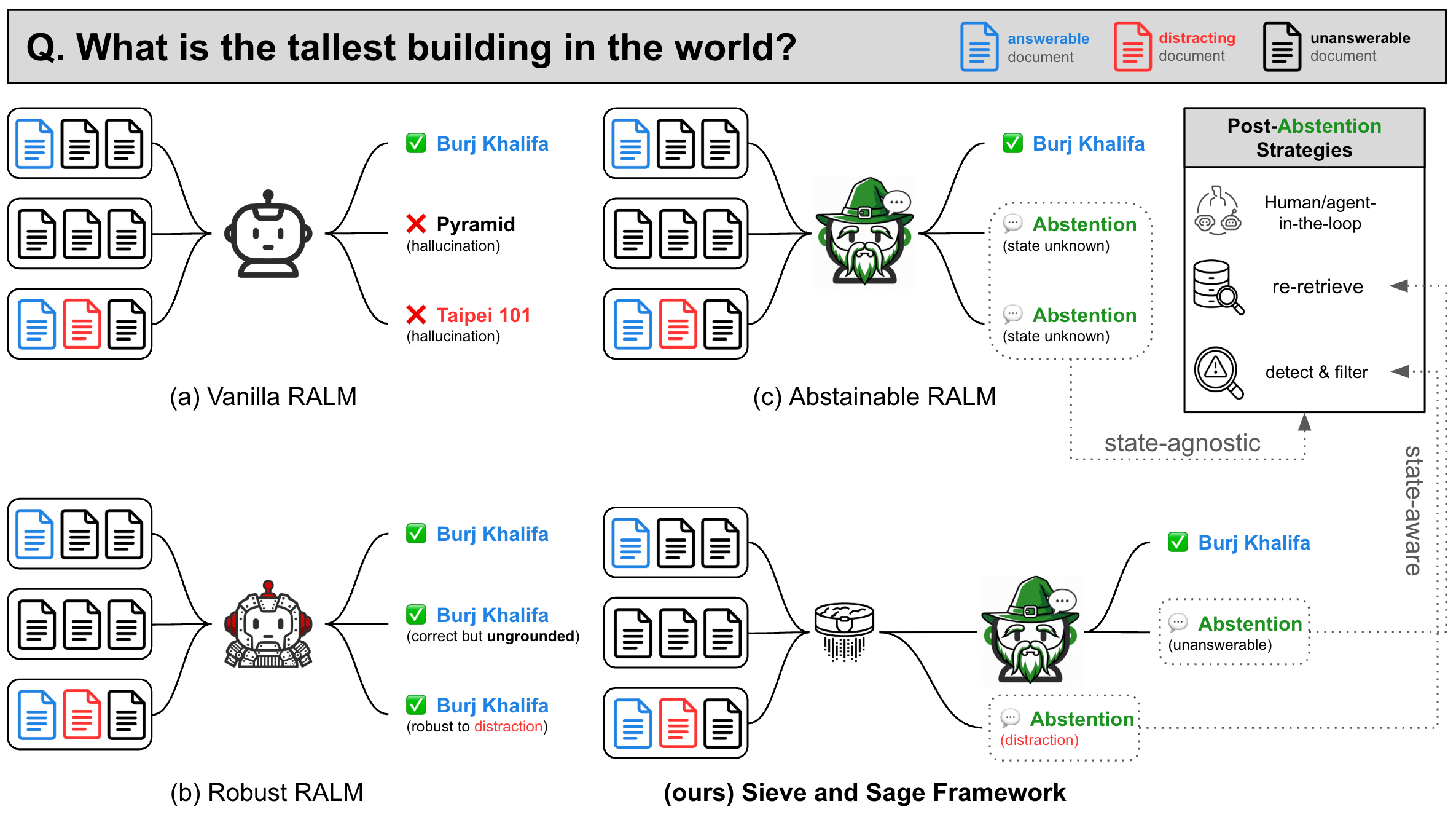}
    \caption{\textbf{Comparison between conventional LLM-centric approaches and Sieve and Sage}. While previous robustness and abstention studies rely solely on the internal judgment of a monolithic LLM, our framework introduces a decoupled architecture: a lightweight Sieve for preemptive distraction filtering and a generative Sage for high-fidelity generation and abstention for unanswerable instances. Unlike (c) conventional abstainable RALMs that produce a state-agnostic abstention signal, (d) our Sieve and Sage framework identifies the underlying retrieval state, enabling state-aware post-abstention strategies such as re-retrieval for unanswerable cases and distraction filtering for distracted cases.}
    \label{fig:sieve-and-sage}
\end{figure*}

\section{Introduction}

\begin{quote}\textit{
Let me perform the Three Sieves test $\cdots $The third sieve is that of utility: If it is not useful, let it not distract my mind.
}
\begin{flushright}---Adapted from a Socratic anecdote \end{flushright}
\end{quote}

Socratic ideas offer a useful metaphor for reliable abstention: filtering distracting information before deliberation (the \textit{Three Sieves test}), and abstaining when the available evidence is insufficient (\textit{knowing what one does not know}). These ideas are particularly relevant to Retrieval-Augmented Language Models (RALMs), whose reliability can degrade when retrieved documents are insufficient, irrelevant, or misleading \cite{yoranMakingRetrievalAugmentedLanguage2024, cuconasuPowerNoiseRedefining2024a}.

For RALMs, the objective is therefore not merely to produce a correct answer, but to provide a response that is faithfully grounded in the retrieved evidence. When the available evidence is insufficient or unreliable, abstention can be preferable to an unsupported answer. This is especially important in high-stakes expert domains, where abstention can itself serve as a useful signal for follow-up actions.

Prior approaches to RALM abstention have largely relied on Large Language Models, either through fine-tuning \cite{sunDivideThenAlignHonestAlignment2025, gul2026mashmodelingabstentionselective} or external LLM-based judges \cite{jorenSufficientContextNew2025a}. By treating retrieval noise as a homogeneous category, such approaches incur substantial computational overhead while providing limited distinction between different causes of abstention.

In contrast, we propose a two-stage framework that distinguishes two forms of retrieval failure: \emph{distracted} states, where relevant evidence is mixed with conflicting or adversarial information, and \emph{unanswerable} states, where sufficient evidence is absent. This distinction operationalizes the two Socratic ideas above: the former calls for screening distracting information before deeper reasoning, whereas the latter calls for recognizing when there is not enough evidence to answer.

Because these failure modes require different handling, we assign them to specialized components. A lightweight BERT-based classifier detects distracted inputs before invoking a costly LLM, while the LLM is reserved for grounded generation and abstention under insufficient evidence. Beyond improving efficiency, identifying the underlying retrieval state also enables state-aware post-abstention strategies tailored to the detected failure mode. Figure~\ref{fig:sieve-and-sage} contrasts this architecture with conventional monolithic RALMs.

\begin{figure*}[t!]
    \centering
    \includegraphics[width=1.0\linewidth]{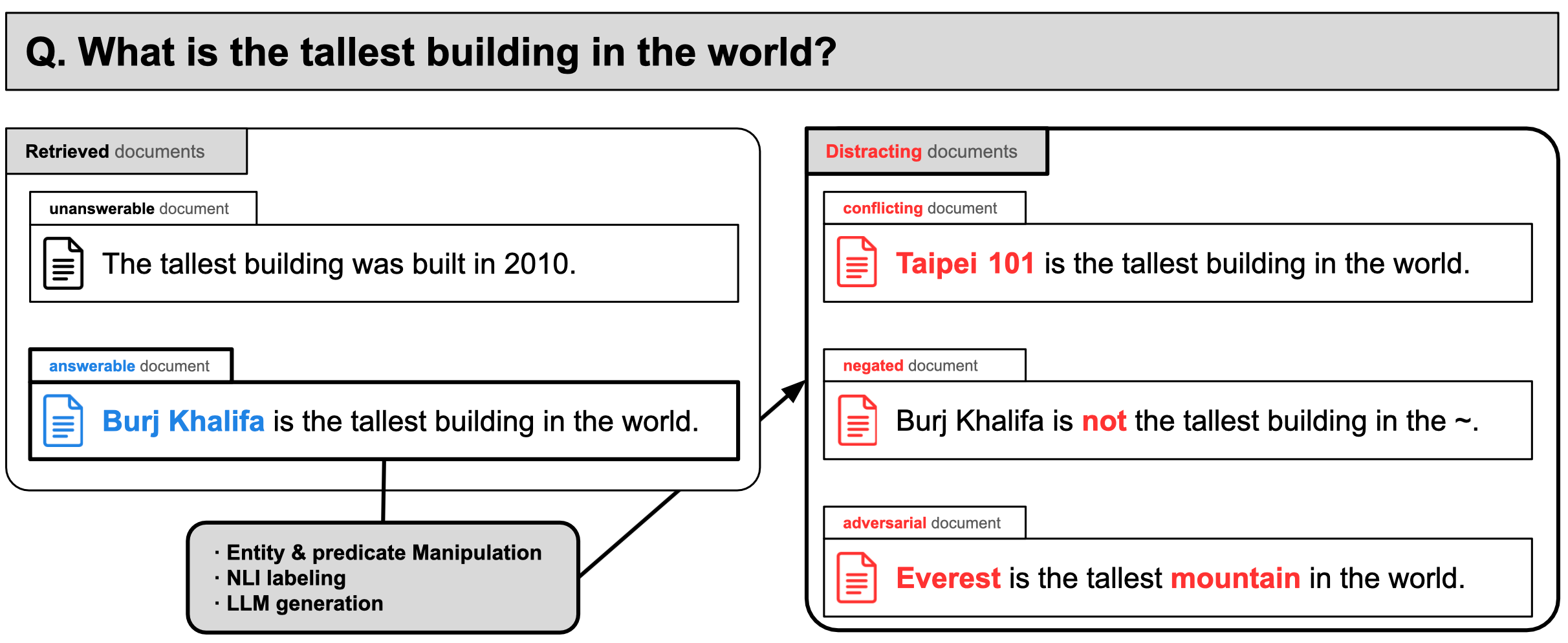}
    \caption{\textbf{Examples of retrieved and simulated documents}. To evaluate the abstention under diverse retrieval noise, we simulated various distracting documents by perturbing answerable documents via entity/predicate manipulation, NLI labeling, and LLM generation.}
    \label{fig:document-example}
\end{figure*}

To evaluate our framework, we systematically simulate sophisticated distracting noise---including conflicting, negated, and adversarial documents---via entity substitution, NLI, and LLM generation. This enables a multi-faceted assessment across both general and expert domains. Our Sieve and Sage framework improves system accuracy by up to 69.4\%p and Macro-F1 by 55.2\%p over vanilla LLMs, while achieving up to a 1.99$\times$ inference speedup over monolithic RALM baselines such as DTA-chatQA. These results demonstrate that explicitly modeling retrieval states can improve both reliability and efficiency.

Our contributions are summarized as follows:
\begin{itemize}
    \item \textbf{Retrieval Noise Decomposition}: We distinguish two forms of retrieval failure---\emph{distraction}, where relevant evidence is mixed with misleading information, and \emph{unanswerability}, where sufficient evidence is absent. This decomposition enables state-aware abstention rather than treating retrieval noise as a single homogeneous category.
    \item \textbf{Efficient Sieve-and-Sage Framework}: We introduce a two-stage architecture that assigns distraction detection to a lightweight Sieve and reserves the generative Sage for grounded answering and abstention under insufficient evidence. This specialization reduces unnecessary LLM computation while exposing the underlying retrieval state for targeted post-abstention handling.
    \item \textbf{Evaluation Across Diverse Domains and Noises}: We evaluate the framework across general and expert-domain datasets under multiple forms of distracting noise. Our approach improves Macro-F1 by up to 55.2\%p while reducing inference latency by up to 1.99$\times$ compared with monolithic baselines.
\end{itemize}

\section{Related Works}

\subsection{RALM Under Retrieval Noise}
Retrieval-Augmented Language Models (RALMs) \cite{ramInContextRetrievalAugmentedLanguage2023a,shiREPLUGRetrievalAugmentedBlackBox2024} ground LLMs in external knowledge for more faithful and reliable responses. However, retrieval can fail due to retriever limitations, corrupted datastores, or adversarial attacks \cite{zhongPoisoningRetrievalCorpora2023, zouPoisonedRAGKnowledgeCorruption2024}. Recent studies therefore investigate RALM behavior under noisy retrieval, where relevant evidence may be absent, irrelevant, or misleading \cite{yoranMakingRetrievalAugmentedLanguage2024, fangEnhancingNoiseRobustness2024, jorenSufficientContextNew2025a}.

Meanwhile, existing evaluations on RALMs have largely focused on general-domain benchmarks such as NQ\cite{leeLatentRetrievalWeakly2019}, WebQ\cite{bordesQuestionAnsweringSubgraph2014}, and HotpotQA 
\cite{yangHotpotQADatasetDiverse2018}. These benchmarks often overlook expert domains, where parametric knowledge may be limited and unsupported responses can be costly \cite{wangDomainRAGChineseBenchmark2024b, thakurKnowingWhenYou2024}. We therefore evaluate our framework across both general and specialized medical, biological, and astronomical domains.

\subsection{Enhanced RALM: Robustness vs. Abstention}
Given that retrieved document sets are inevitably noisy, prior work has largely pursued two complementary directions: \textbf{robustness} and \textbf{abstention}.

\paragraph{Robustness in RALMs}
Robustness-oriented approaches aim to generate correct answers despite irrelevant or distracting retrieved contexts.
Existing strategies include exposing models to low-ranked contexts during supervised fine-tuning \cite{yoranMakingRetrievalAugmentedLanguage2024}, adaptive adversarial training \cite{fangEnhancingNoiseRobustness2024}, and auxiliary lightweight modules for assessing or filtering retrieved evidence \cite{yanCorrectiveRetrievalAugmented2024, kimRERAGImprovingOpenDomain2024a, yuChainofNoteEnhancingRobustness2024}.

However, robustness can be misaligned with the practical goals of RALMs: by forcing an answer, such models may compromise faithfulness or, more critically, fail to signal uncertainty. In expert domains, the absence of an explicit error signal can be particularly dangerous, as a plausible but ungrounded response may be more harmful than an admission of uncertainty.

\paragraph{Abstention in RALMs}
Abstention-oriented approaches instead enable RALMs to refrain from answering when the available evidence is insufficient or unreliable. Existing approaches include post-training models to learn abstention behavior \cite{sunDivideThenAlignHonestAlignment2025,gul2026mashmodelingabstentionselective}, prompting LLMs to reason about retrieval reliability through in-context learning 
\cite{parkEnhancingRobustnessRetrievalAugmented},
and using external LLM-based sufficiency judges 
\cite{jorenSufficientContextNew2025a}.

However, previous abstention approaches remain largely LLM-centric, incurring substantial computational costs while remaining vulnerable to the over-trust issue, where LLMs may blindly favor their own generations over a more reliable silence \cite{liThinkTwiceTrusting2024}.

In contrast, we decompose retrieval noise into two distinct states---\emph{unanswerable} and \emph{distracted} ($S_\text{unans}$ and $S_\text{distract}$)---and assign them to different stages of the abstention pipeline. This decomposition enables a lightweight classifier to preemptively detect distracting evidence before LLM inference, reducing computational overhead while supporting early exit and state-specific post-abstention strategies.

Most closely related to our work, \citet{hong-etal-2024-gullible} employ a lightweight discriminator to detect counterfactual evidence for robust answer generation. While their focus is robustness under conflicting evidence, our framework targets system-level abstention by explicitly distinguishing distraction from missing evidence and handling these states separately.

\begin{figure*}[t!] 
    \centering
    \begin{subfigure}[b]{0.48\textwidth}
        \centering
        \includegraphics[width=\textwidth]{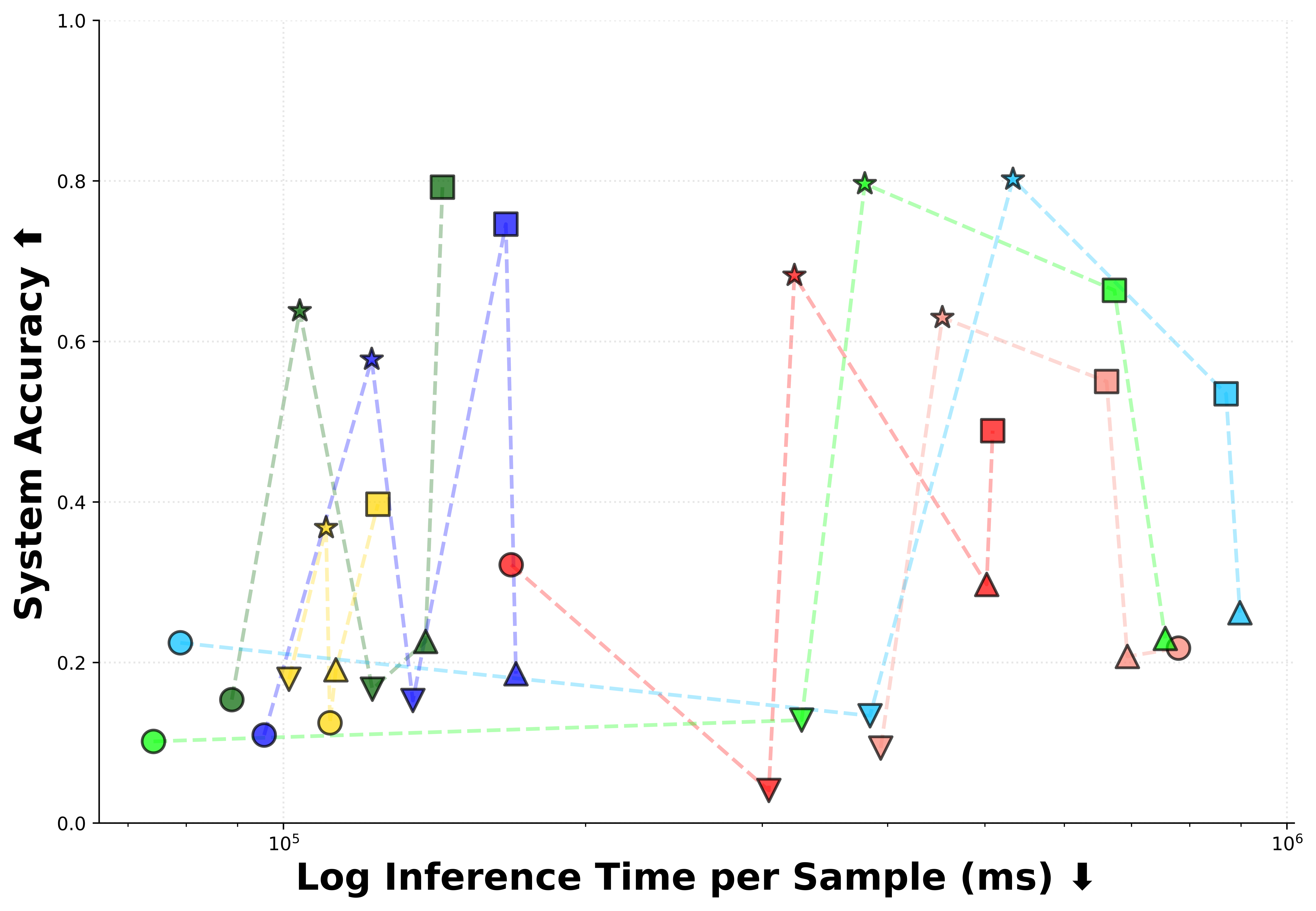}
        \caption{Inference time vs. System accuracy}
        \label{fig:pareto_system-accuracy}
    \end{subfigure}
    \hfill 
    \begin{subfigure}[b]{0.48\textwidth}
        \centering
        \includegraphics[width=\textwidth]{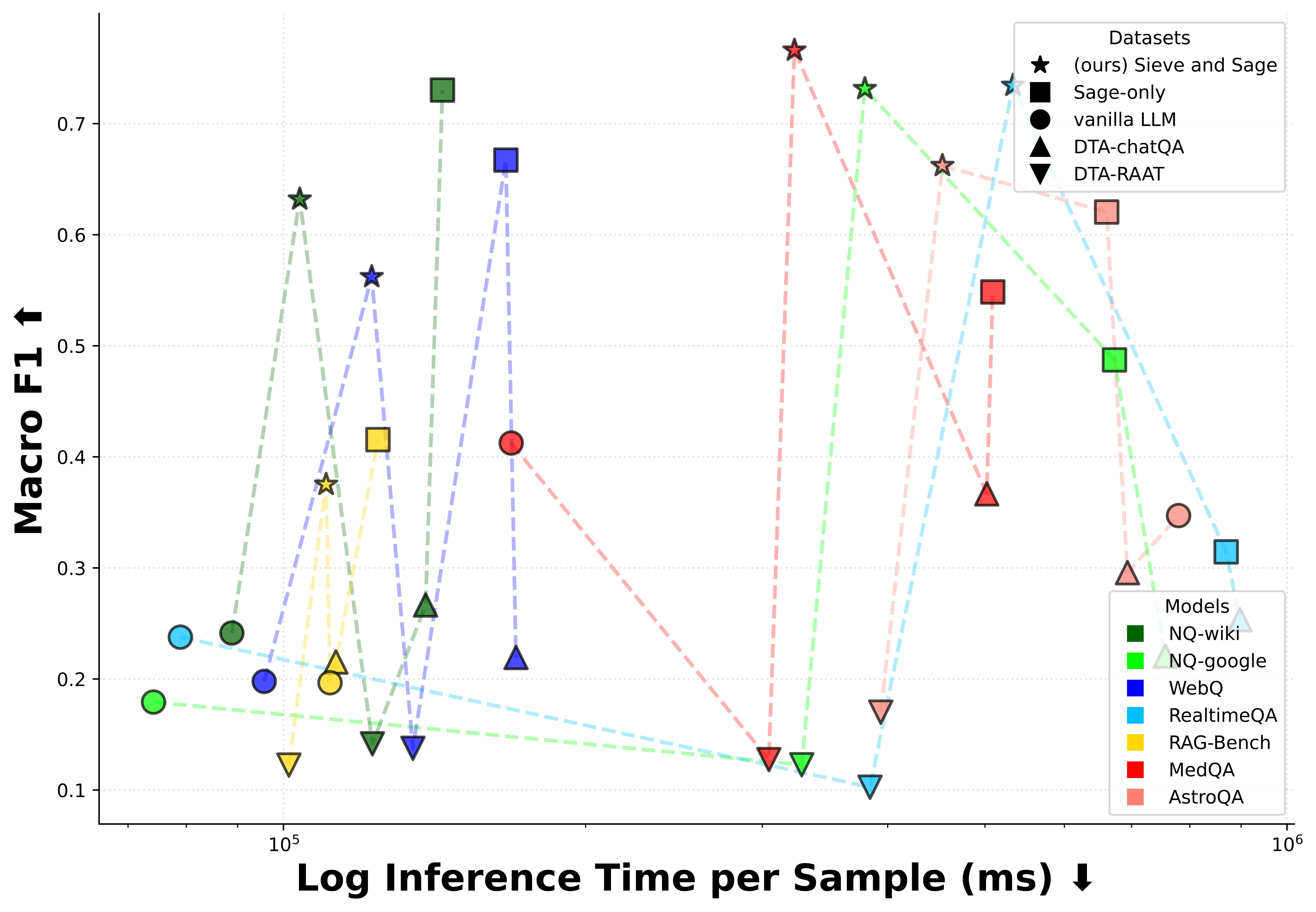}
        \caption{Inference time vs. Macro F1}
        \label{fig:pareto_macro-f1}
    \end{subfigure}
    
    \caption{\textbf{Pareto graph of inference time versus System Accuracy/Macro F1}. Our framework($\star$) is up to 2$\times$ faster than other LLM-centric approaches with comparable performance, especially achieving the highest F1 scores in expert domain datasets.}
    \label{fig:pareto_main}
\end{figure*}

\section{Methodology}

\subsection{Problem Formulation} \label{ssec:problem formulation}
To formalize RALM with abstention, we characterize the retrieval state $S$ of the document set $D = \{d_1, d_2, \dots, d_k\}$ retrieved for a given query $q$. Prior work on RALM robustness and abstention has often treated unreliable retrieval as a coarse category without explicitly distinguishing its underlying causes \cite{yoranMakingRetrievalAugmentedLanguage2024, xiangCertifiablyRobustRAG2024a}. We instead propose a more fine-grained tripartite taxonomy ($S \in \{S_{\text{ans}}, S_{\text{unans}}, S_{\text{distract}}\}$) based on document-level evidence composition. 

This distinction further enables state-specific follow-up strategies after abstention, as discussed in Section~\ref{sec:post-abstention-strategies}.

\paragraph{Document-level Evidence Assessment} 
We categorize each document $d \in D$ by lexical matching (checking if $d$ contains the ground-truth answer $a^*$) and semantic consistency via NLI against a reference sentence generated by LLM conditioned on $q$ and $a^*$:

\begin{itemize}
    \item \textbf{Answerable document} ($d_\text{ans}$): Contains $a^*$ and entails the reference sentence. It serves as primary evidence.
    
    \item \textbf{Unanswerable document} ($d_\text{unans}$): Neither contains $a^*$ nor entails the reference. It provides no valid grounds, while not misleading the system.
    
    \item \textbf{Distracting document} ($d_\text{distract}$): Contains misleading or inconsistent evidence that can divert the system from the grounded answer. In our controlled evaluation, distracting documents are instantiated as three types---\emph{conflict}, \emph{negated}, and \emph{adversarial} documents---by manipulating key entities or their associated predicates. Although all three types induce the same set-level distracted state ($S_{\text{distract}}$), they represent diverse forms of retrieval noise, including factual inconsistency, logical negation, and targeted misleading evidence. Importantly, the Sieve is trained using only conflict documents as positive distracted examples and is evaluated across all three types, allowing us to assess its generalization beyond the type of noise observed during training. Detailed implementation is provided in Section~\ref{ssec:dataset construction} and Figure~\ref{fig:document-example}.
\end{itemize}

Documents that do not meet these criteria---such as those containing $a^*$ but remaining semantically neutral to the reference sentence---are marked as \textbf{neutral} ($d_\text{neutral}$) and assumed to have a negligible impact on the retrieval state $S$. We empirically validate this assumption in Appendix~\ref{app:neutral-ablation}, where adding up to three neutral documents results in minor performance variations.

\paragraph{Set-level Retrieval States}
Building on the document-level assessment, we define three retrieval states: 

\begin{itemize}
    \item \textbf{Answerable state} ($S_\text{ans}$): $D$ contains at least one $d_\text{ans}$ and no $d_\text{distract}$. It may include $d_\text{unans}$ or $d_\text{neutral}$ as noise. The system is expected to generate $a^*$.
    
    \item \textbf{Unanswerable state} ($S_\text{unans}$): $D$ consists entirely of $d_\text{unans}$. The system should recognize the lack of evidence and refrain from answering ($a_\text{unans}$). 

    \item \textbf{Distracted state} ($S_{\text{distract}}$):
    $D$ contains at least one $d_{\text{ans}}$ and at least one
    $d_{\text{distract}}$. The system should detect the
    evidentiary distraction and reject the set
    ($a_{\text{distract}}$).
\end{itemize}

Under this taxonomy, RALM with abstention is represented by the decision function $f$:
$$\resizebox{0.98\columnwidth}{!}{$
\begin{aligned}
f(q,D)=
\begin{cases}
a_{\text{grounded}}, & \text{if } D \in S_{\text{ans}},\\
a_{\text{unans}}, & \text{if } D \in S_{\text{unans}},\\
a_{\text{distract}}, & \text{if } D \in S_{\text{distract}}.
\end{cases}
\end{aligned}
$}$$
where $a_{\text{grounded}}$ denotes a grounded answer generated from the retrieved evidence, while $a_{\text{unans}}$ and $a_{\text{distract}}$ denote distinct abstention signals indicating insufficient and distracting evidence, respectively.

\subsection{Sieve and Sage: Efficient Two-stage Abstention Framework} \label{ssec:sieve-and-sage}
To implement $f(q, D)$ defined above, we propose a two-stage framework that decouples distraction detection from grounded generation by employing \textbf{Sieve} (a lightweight classifier) and \textbf{Sage} (a generative LLM). Our framework separates three functions that are often conflated in monolithic RALMs: (1) distraction detection by the Sieve, (2) missing-evidence detection by the Sage, and (3) grounded answer generation by the Sage. Abstention is then triggered when either distraction or missing evidence is identified.

\paragraph{Stage 1: The Sieve for Distraction Detection} 
The Sieve $g_{\phi}$ acts as an efficient gatekeeper to preemptively identify distracted document sets ($S_\text{distract}$). Intercepting distracting inputs before LLM invocation reduces unnecessary generation and computational overhead. Formally:
$$
\hat{y}_\text{distract} = \mathbb{I}(\sigma (g_{\phi}(q, D)) \geq \tau_{distract})
$$
Given a predefined threshold $\tau_{\text{distract}}$, if $\hat{y}_{\text{distract}}=1$, the system immediately outputs $a_{\text{distract}}$, bypassing the subsequent LLM invocation entirely. We set $\tau_{\text{distract}}=0.5$ as the default threshold, while the Sieve's classification performance remains stable across a practical range of alternative thresholds (Appendix~\ref{app:threshold-sensitivity}).

\begin{table*}[t!]
\centering
\resizebox{\textwidth}{!}{%
\begin{tabular}{@{}lrrrrrrrr@{}}
\toprule
                     & \multicolumn{5}{c}{\textbf{General domain}}                         & \multicolumn{3}{c}{\textbf{Expert domain}}            \\ \cmidrule(lr){2-6}  \cmidrule(lr){7-9}
test(train) datasets & NQ-wiki       & NQ-google    & WebQ      & RealtimeQA   & RAG-Bench & MedQA            & BioASQ           & AstroQA         \\ \midrule
document source      & Wikipedia     & Google search & Wikipedia & Google search & Wikipedia & PubMed abstracts & PubMed abstracts & Astro abstracts \\
document length      & 165.93        & 152.29       & 172.63    & 145.27       & 409.29    & 1024.47          & 955.77           & 1046.2          \\ \midrule
\# total samples      & 3297 (44030)  & 226 (2182)   & 2096      & 187          & 4052      & 491 (2087)       & 610 (3997)       & 289 (1489)      \\ \midrule
\# answerable         & 889 (15086)\textsuperscript{*} & 51 (418)\textsuperscript{*}   & 545\textsuperscript{*}     & 34\textsuperscript{*}         & 912\textsuperscript{*}     & 116 (409)        & 75 (536)\textsuperscript{*}       & 99 (460)        \\
\# unanswerable       & 483 (20688)\textsuperscript{*} & 49 (382)\textsuperscript{*}   & 296\textsuperscript{*}     & 66\textsuperscript{*}         & 419\textsuperscript{*}     & 159 (539)        & 397 (3236)\textsuperscript{*}     & 86 (409)        \\
\# conflict           & 530 (8256)\textsuperscript{*}  & 47 (396)     & 360\textsuperscript{*}     & 34           & 907\textsuperscript{*}     & 70 (273)         & 33 (225)         & 47 (216)        \\
\# negated            & 889           & 51           & 545       & 34           & 907       & 77               & 59               & 47              \\
\# adversarial        & 506           & 28           & 350       & 19           & 907\textsuperscript{*}     & 69               & 46               & 10              \\ \bottomrule
\end{tabular}%
}
\caption{\textbf{Statistics of test (and train) datasets.} An asterisk (*) indicates that the documents are adopted from prior studies rather than retrieved or simulated in this work.}
\label{tab:datasets-stats}
\end{table*}

\paragraph{Stage 2: The Sage for Grounded Generation}
Document sets that pass the Sieve (i.e., $\hat{y}_{\text{distract}} = 0$) are forwarded to the Sage, a generative LLM $P_{\theta}$. Having been screened for distracting information, the Sage focuses on either generating a grounded answer or identifying the absence of sufficient evidence ($S_{\text{unans}}$). Formally,

$$\resizebox{0.98\columnwidth}{!}{$
\begin{aligned}
    \hat{a} \sim P_{\theta}(a \mid q, D), \qquad
    \hat{a} =
    \begin{cases}
    a_{\text{grounded}}, & \text{if }D \in S_{\text{ans}},\\
    a_{\text{unans}}, & \text{if }D \in S_{\text{unans}}.
    \end{cases}
\end{aligned}
$}
$$
where the Sage is fine-tuned to generate a grounded answer \(a_{\text{grounded}}\) for \(S_{\text{ans}}\) and an abstention signal \(a_{\text{unans}}\) for \(S_{\text{unans}}\). Unlike the Sieve, the Sage does not rely on an explicit decision threshold; it directly generates either a grounded answer or the designated abstention signal.

This hierarchical design ensures that the high-cost LLM is invoked only after the retrieved document set has been screened for distraction by the lightweight Sieve.

\paragraph{Fine-tuning Strategy}
To optimize the two-stage framework, we employ task-specific optimization for each component. During training, the Sieve learns a binary decision boundary between $S_{\text{ans}}$ and  $S_{\text{conflict}}$:
$$
\mathcal{L}_{\text{Sieve}} = -\sum_{i} \left[ y_i \log(p_i) + (1 - y_i) \log(1 - p_i) \right],
$$

where $p_i = \sigma(g_{\phi}(q, D_i))$.

In contrast, the Sage is optimized via supervised fine-tuning (SFT) to maximize the likelihood of the ground-truth $a^*$ for $S_\text{ans}$ and the designated abstention signal $a_\text{unans}$ for $S_\text{unans}$:
$$
\mathcal{L}_{\text{Sage}} = -\sum_{t=1}^{|Y|} \log P_{\theta}(y_t \mid y_{<t}, q, D),
$$

These task-specific objectives enable the Sieve to specialize in distraction detection while training the Sage to generate grounded answers when sufficient evidence is available and abstain when the evidence is insufficient.

\section{Experimental Setup}

\subsection{Dataset Construction} \label{ssec:dataset construction}
To evaluate the performance of RALMs across diverse contexts, we employed 5 general-domain datasets (NQ-wiki\cite{parkEnhancingRobustnessRetrievalAugmented}, NQ-google\cite{xiangCertifiablyRobustRAG2024a},  WebQ\cite{parkEnhancingRobustnessRetrievalAugmented}, RealtimeQA\cite{xiangCertifiablyRobustRAG2024a}, RAG-Bench\cite{fangEnhancingNoiseRobustness2024}) and 3 expert-domain datasets (AstroQA\cite{haanAstroMLab3Achieving2025}, MedQA\cite{palMedMCQALargescaleMultiSubject2022}, BioASQ\cite{nentidis2025overviewbioasq2025thirteenth}).

To maintain consistency with established benchmarks, we adopt the document sets and retrieval noise from prior studies for the datasets marked with an asterisk (*) in Table~\ref{tab:datasets-stats}.

For BioASQ and AstroQA datasets, we utilized ColBERTv2 \cite{santhanamColBERTv2EffectiveEfficient2022a} to retrieve the top-20 candidate documents per query. Following the criteria defined in Section~\ref{ssec:problem formulation}, documents are labeled using domain-specific NLI models\footnote{We used \href{https://huggingface.co/cnut1648/biolinkbert-mednli}{cnut1648/biolinkbert-mednli} for MedQA and BioASQ, 
\href{https://huggingface.co/kauffinger/scibert_scivocab_uncased-mnli}{auffinger/scibert\_scivocab\_uncased-mnli} for AstroQA, 
and \href{https://huggingface.co/MoritzLaurer/mDeBERTa-v3-base-xnli-multilingual-nli-2mil7}{MoritzLaurer/mDeBERTa-v3-base-xnli-multilingual-nli-2mil7} for other general datasets.} and the retrieval state $S$ is determined by the top-5 documents. An ablation over the number of documents $K$ is provided in Appendix~\ref{app:k-ablation}.

\paragraph{Simulating distracting documents $d_\text{distract}$}
To evaluate abstention under diverse retrieval noise, we manipulate reference answer sentences into three distracting types ($d_\text{conflict}$, $d_\text{negated}$, $d_\text{adversarial}$). Each is expanded into a full-length document matching the data store's length distribution. The manipulation is validated via entity substitution, NLI, and LLM verification, following and extending prior works \cite{parkEnhancingRobustnessRetrievalAugmented, fangEnhancingNoiseRobustness2024}:

\begin{itemize}
    \item \textbf{Conflicting document} ($d_\text{conflict}$): We replace the ground-truth entity $a^*$ with a plausible but incorrect alternative. Validation ensures $a^*$ is absent, and the document explicitly contradicts the reference.
    
    \item \textbf{Negated document} ($d_{\text{negated}}$): We retain $a^*$ but negate its associated predicate. Validation ensures $a^*$ remains while the semantic meaning is contradictory, thereby testing the model's reliance on keyword heuristics.
    
    \item \textbf{Adversarial} ($d_{\text{adversarial}}$): 
    Both the entity and its context are shifted to a different subject. Validation ensures the document neither contains the ground-truth entity $a^*$ nor is contradictory to the reference.
\end{itemize}

Concrete examples and analyses of the simulated documents are provided in Figure~\ref{fig:document-example} and Appendix~\ref{sec:validation-of-simulated-distracting-documents}.

\begin{figure*}[t!]
    \begin{subfigure}[b]{0.48\textwidth}
        \centering
        \includegraphics[width=1.0\linewidth]{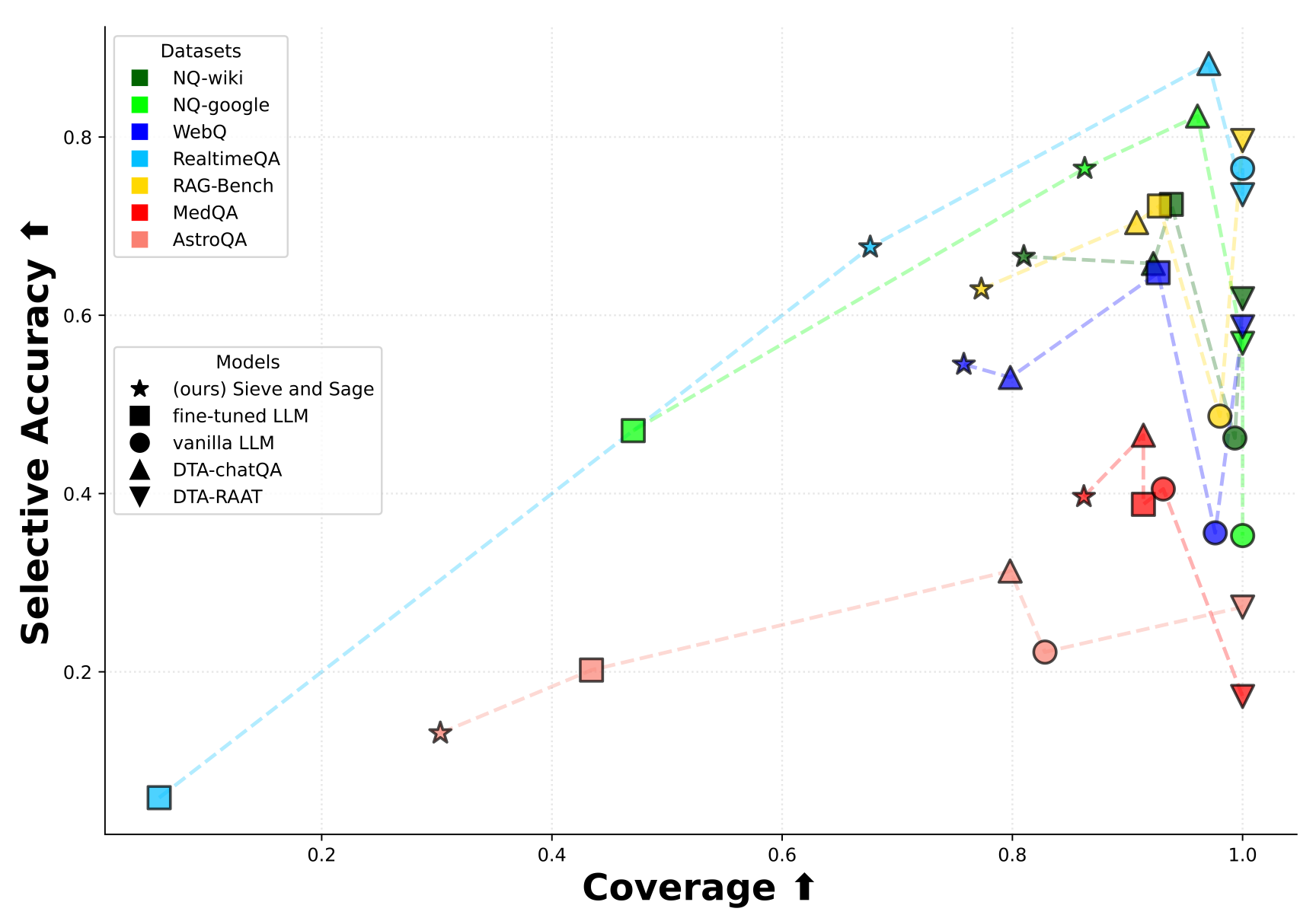}
        \caption[Coverage vs. Selective Accuracy]{\textbf{Coverage vs. Selective Accuracy \\ Under Answerable Retrieval State}}
        \label{fig:calibration}
    \end{subfigure}
\hfill 
    \begin{subfigure}[b]{0.48\textwidth}
        \centering
        \includegraphics[width=1.0\linewidth]{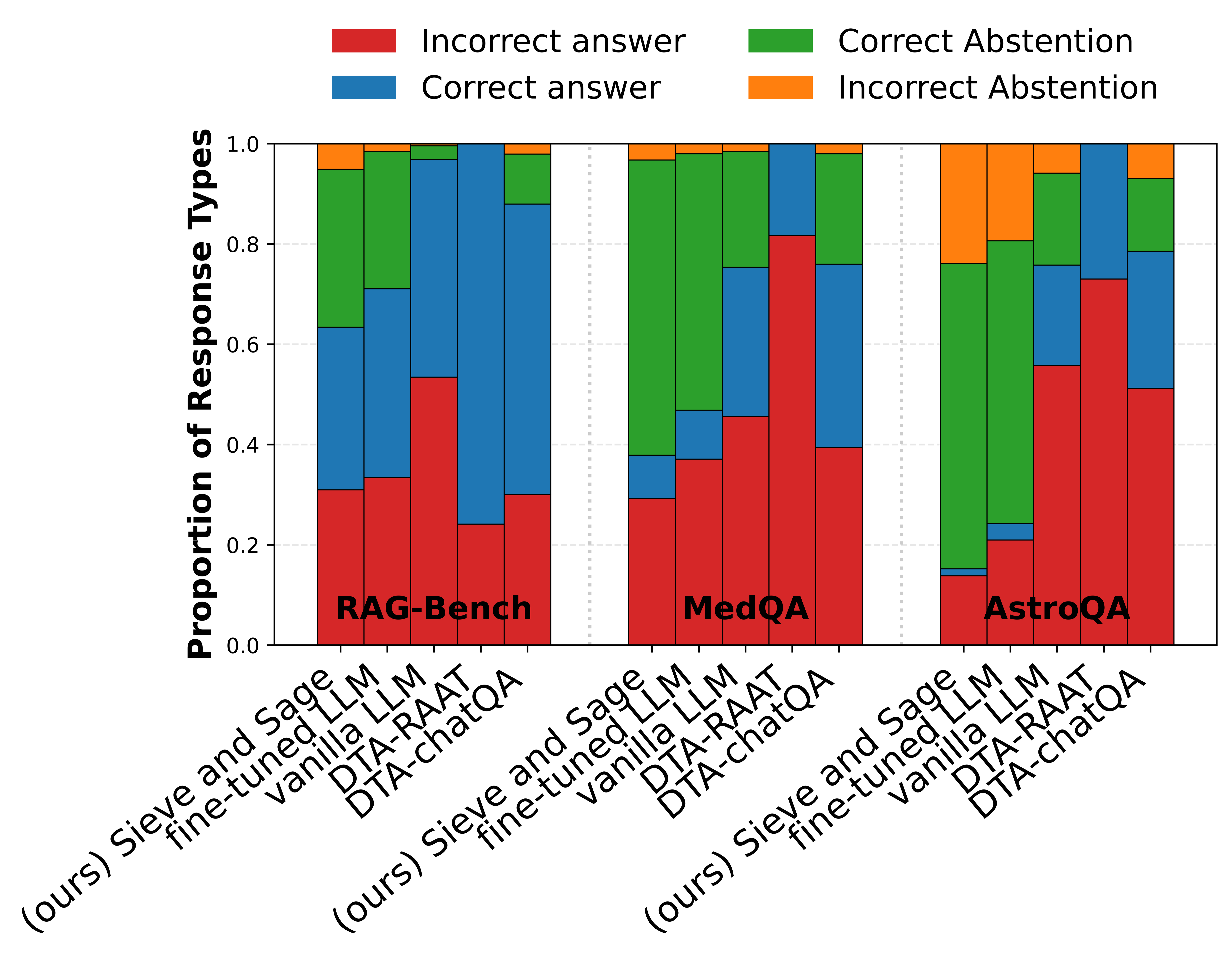}
        \caption[Distribution of Model Response]{\textbf{Distribution of Model Response \\ Regardless of the Retrieval States}} 
        \label{fig:reliability}
    \end{subfigure}
\caption{\textbf{Selective Prediction and System Reliability}. We evaluate system reliability through two complementary analyses: (a) Coverage and Selective Accuracy for answerable examples (precision on $S_\text{ans}$) and (b) distribution of model response across all retrieval states. Although Sieve and Sage operates at lower coverage than more aggressive baselines, this reflects a more conservative response policy and is accompanied by fewer incorrect responses on several datasets.}
\label{fig:calibration_and_reliability}
\end{figure*}

\subsection{Implementation Details}
For our framework, we use Longformer \cite{beltagyLongformerLongDocumentTransformer2020}\footnote{\href{https://huggingface.co/allenai/longformer-base-4096}{allenai/longformer-base-4096}} as the Sieve and Llama-3-8B-Instruct \cite{grattafioriLlama3Herd2024a}\footnote{\href{https://huggingface.co/meta-llama/Meta-Llama-3-8B-Instruct}{meta-llama/Meta-Llama-3-8B-Instruct}} as the Sage. We choose Longformer to accommodate long retrieved contexts, particularly in expert-domain datasets. See Appendix~\ref{app:extended-results} for results with alternative Sieve backbones and Sage model families.

Each component is fine-tuned on NQ dataset \cite{leeLatentRetrievalWeakly2019} according to the strategies described in Section~\ref{ssec:sieve-and-sage}. Detailed configurations are provided in Appendix~\ref{sec:technical-specifications}. A detailed comparison of the computational footprint of the two components, including parameter count, VRAM usage, and inference FLOPs, is provided in Appendix~\ref{app:computational-cost}.

\paragraph{Domain Adaptation}
To evaluate our framework in specialized domains, we adapt both the generator and the classifier.

For LLM generation, expert domains require strong parametric knowledge. We therefore employ domain-specific variants: \texttt{Med42-8B} \cite{christopheMed42EvaluatingFineTuning2024a} for MedQA and BioASQ, and \texttt{AstroSage-8B} \cite{haanAstroMLab3Achieving2025} for AstroQA.

In contrast, the classifier primarily detects distractions, relying more on structural reasoning than domain-specific knowledge. To accommodate the distributional characteristics of expert documents (e.g., longer passages and terminology), we perform minimal structural adaptation using fewer than 200 samples per domain. Standalone evaluations further show that the Sieve achieves substantial adaptation gains with fewer than 200 target-domain examples and transfers effectively across expert domains (Appendix~\ref{sec:domain-adaptation-for-the-sieve}).

\subsection{Evaluation Metrics}

We evaluate the performance of Sieve and Sage along two primary dimensions: response accuracy for answerable instances and retrieval state classification.

For response accuracy, we measure how often the model attempts to answer answerable queries (\textbf{Coverage}) and how often those responses are correct (\textbf{Selective Accuracy}). Coverage denotes the proportion of answerable instances for which the system chooses to respond, while Selective Accuracy measures the correctness of the responses conditioned on those instances where the system produces an answer.

For classification, we treat the decision process as a tripartite task ($S_\text{ans}$, $S_\text{unans}$, $S_\text{distract}$) and report \textbf{State-Specific F1} scores (Answerable, Unanswerable, and Distract F1) to evaluate the model's ability to identify different retrieval states. In addition, we report \textbf{Macro-F1} (unweighted mean) to provide a balanced assessment across all categories and prevent bias toward the majority state.

Finally, we report \textbf{System Accuracy} to jointly evaluate response generation and state classification. A prediction is considered correct if the system either produces the ground-truth answer $a^*$ for $S_\text{ans}$ or appropriately abstains for $S_\text{unans}$ and $S_\text{distract}$.

\subsection{Baselines}
We compare our framework against three other approaches: (1) vanilla LLM, (2) LLM fine-tuned to abstain when the documents are noisy (\textbf{Sage-only}) \cite{zhangRAFTAdaptingLanguage2024}, and (3) RALM aligned with knowledge-quadrant preference data to enable abstention (\textbf{DTA-chatQA}, \textbf{DTA-RAAT}) \cite{sunDivideThenAlignHonestAlignment2025}. Compared to DTA-chatQA, DTA-RAAT is trained with a stronger emphasis on robustness to retrieval noise, aiming to generate correct answers even in the presence of noisy or distracting documents.

\begin{figure*}[t!]
\centering
\begin{subfigure}{0.23\linewidth}
    \centering
    \includegraphics[width=\linewidth]{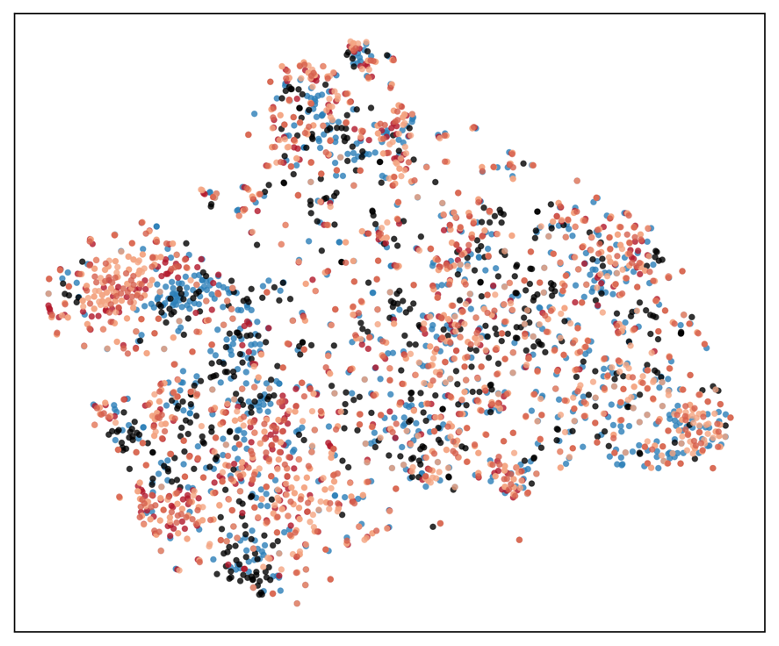}
    \caption{Vanilla \\    \texttt{Longformer}}
    \label{fig:tsne_vanilla}
\end{subfigure}
\hfill
\begin{subfigure}{0.23\linewidth}
    \centering
    \includegraphics[width=\linewidth]{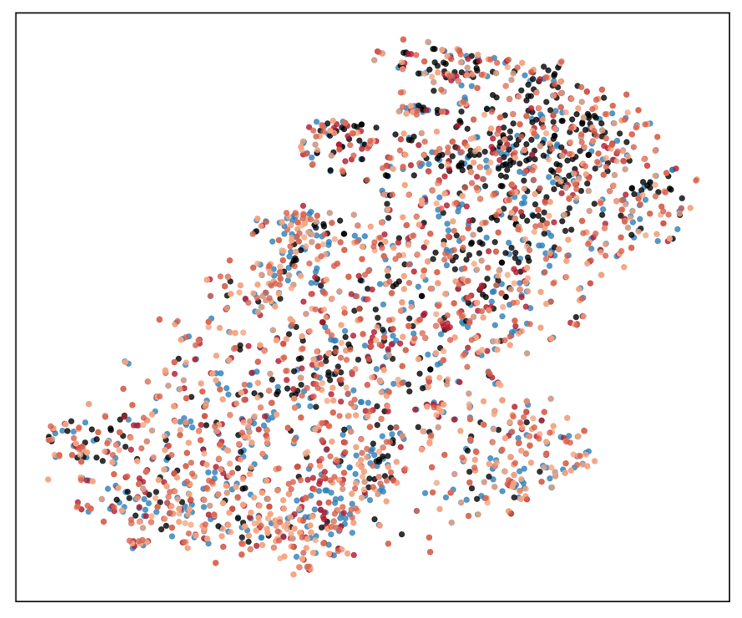}
    \caption{Binary: \\ $S_{\text{ans}}$ vs.\ $S_{\text{unans}}$}
    \label{fig:tsne_unanswerable}
\end{subfigure}
\hfill
\begin{subfigure}{0.23\linewidth}
    \centering
    \includegraphics[width=\linewidth]{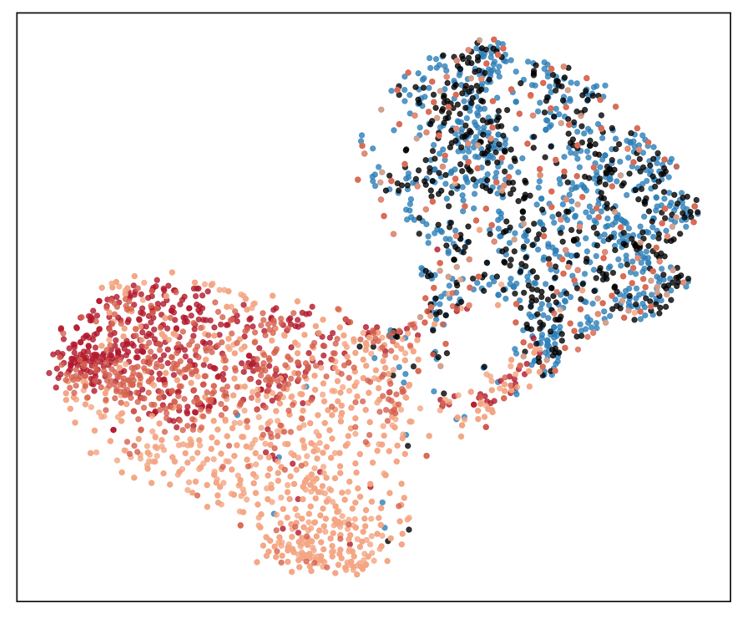}
    \caption{Binary: \\ $S_{\text{ans}}$ vs.\ $S_{\text{conflict}}$}
    \label{fig:tsne_conflict}
\end{subfigure}
\hfill
\begin{subfigure}{0.23\linewidth}
    \centering
    \includegraphics[width=\linewidth]{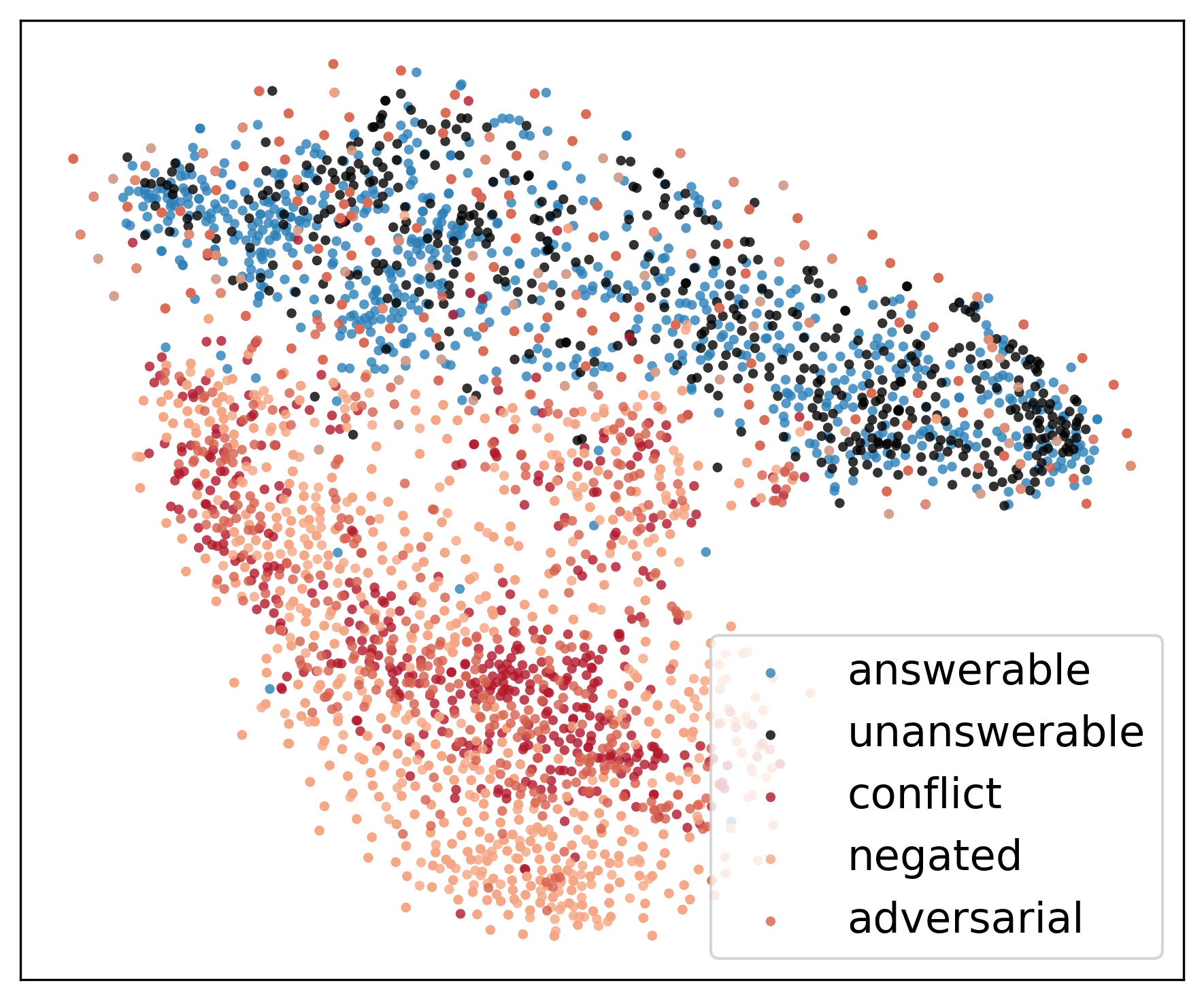}
    \caption{Three-way: $S_{\text{ans}}$,\\     $S_{\text{unans}}$, and $S_{\text{conflict}}$}
    \label{fig:tsne_unanswerable_conflict}
\end{subfigure}
\caption{\textbf{t-SNE comparison of Sieve training objectives}. Our model (c), fine-tuned to classify $S_\text{conflict}$ from $S_\text{ans}$, achieves a more distinct separation of distracting noise compared to the model (d) fine-tuned to classify answerable, unanswerable, and conflicting states at once. This validates that a specialized focus on factual consistency establishes a precise decision boundary for identifying retrieval distractions.}
\label{fig:tsne_all}
\end{figure*}

\section{Experimental Results and Analysis}

\subsection{Efficiency and Effectiveness Analysis}
Figure~\ref{fig:pareto_main} illustrates the Pareto graph for System Accuracy and Macro F1 against average inference latency. Across all datasets, Sieve and Sage achieve competitive performance while maintaining lower latency than most monolithic LLM baselines. Notably, compared to DTA-chatQA, our framework achieves up to a 1.99$\times$ speedup (379.6 ms vs. 756.3 ms on NQ-google) while delivering a 56.6\%p gain in System Accuracy (0.796 vs. 0.230).

The advantages of our decoupled architecture are most pronounced in expert domains. While monolithic LLMs suffer from increased latency and severe performance degradation when processing complex, distracting documents, our framework maintains stability. For instance, on MedQA, Sieve and Sage achieve the highest System Accuracy (0.682) and Macro F1 (0.766) compared to DTA-chatQA (0.297, 0.367), while simultaneously reducing inference time by over 35\% (322.9 ms vs. 502.1 ms). The efficiency gain is further supported by the Sieve's substantially smaller computational footprint---approximately 54$\times$ fewer parameters and 14$\times$ fewer inference FLOPs than the Sage---which amortizes the cost of expensive LLM inference by avoiding unnecessary Sage calls (Appendix~\ref{app:computational-cost}).

These empirical results confirm that explicitly decoupling distraction detection from core reasoning effectively mitigates computational overhead, establishing an optimal balance in both speed and reliability. Comprehensive numerical results and robustness analyses across alternative retrievers, Sieve backbones, and Sage model families are provided in Appendix~\ref{app:extended-results}.

\subsection{Reliability Analysis}
Figure~\ref{fig:calibration} illustrates the relationship between Coverage and Selective Accuracy for answerable instances, while Figure~\ref{fig:reliability} evaluates overall system reliability by analyzing the distribution of model response regardless of the retrieval states.

A key observation is that Sieve and Sage intentionally sacrifice a small amount of Coverage and Selective Accuracy to significantly reduce incorrect answers. On MedQA, Coverage and Selective Accuracy decrease from 0.914 and 0.509 to 0.862 and 0.460, respectively, compared to DTA-chatQA. However, incorrect answers drop by 56.1\% (319 $\to$ 140), while correct abstentions increase substantially (108 $\to$ 289).

In contrast, baseline models such as the Vanilla LLM and DTA-RAAT attempt to respond to almost every query (Coverage $\simeq$ 1.0). This aggressive strategy leads to a large accumulation of incorrect responses. On MedQA, the Vanilla LLM produces 323 incorrect answers among 370 responses (87.3\%), while on AstroQA, DTA-RAAT produces 262 incorrect answers among 289 responses (90.7\%).

These results highlight a key limitation of existing RALMs: maximizing coverage in noisy environments leads to many incorrect responses. In contrast, Sieve and Sage demonstrates that a calibrated, conservative generation strategy is crucial for reliable deployment, especially in high-stakes expert domains.

\subsection{Analysis of Sieve Design}
Figure~\ref{fig:tsne_all} shows the t-SNE visualization of CLS token embeddings for the Sieve under different training objectives. Our proposed approach (c), which focuses on distinguishing the conflict state from the answerable state, exhibits the most distinct separation between clusters compared to other variants.

In contrast, the model (d), fine-tuned to classify all retrieval states ($S_\text{ans}$, $S_\text{unans}$, $S_\text{conflict}$) simultaneously, produces substantial overlap in the embedding space. This suggests that the multi-class objective weakens the model’s ability to detect subtle factual inconsistencies.

Interestingly, although the Sieve was trained only on answerable and conflicting examples, it generalizes well to unseen distracted states such as negated or adversarial cases. This indicates that the Sieve does not simply memorize specific noise patterns, but instead learns to detect underlying inconsistencies among the documents.

An isolated component ablation further confirms that both the Sieve and Sage contribute independently to the overall performance, with their combination yielding the strongest results (Appendix~\ref{app:component-ablation}).

\section{State-Aware Post-Abstention Strategies}
\label{sec:post-abstention-strategies}

When a RALM abstains, the outcome need not be treated as a terminal failure. Instead, the abstention signal can trigger post-abstention strategies, such as additional reasoning or self-consistency. Sieve and Sage further identifies the underlying cause of abstention---missing evidence ($S_{\text{unans}}$) or distracting evidence ($S_{\text{distract}}$)---enabling state-specific follow-up actions as follows. 

\paragraph{Human/Agent-in-the-loop}
Regardless of the underlying abstention state, abstained queries can be escalated to a human expert or a more capable LLM. Such a cascade reserves expensive expert resources for the subset of examples identified as unreliable by the base system.

\paragraph{Dynamic Re-retrieval}
When the Sage identifies an unanswerable state ($S_{\text{unans}}$), the system can initiate an additional retrieval step to search for missing evidence. This directly addresses the underlying cause of abstention rather than repeatedly reasoning over an insufficient document set.

\paragraph{Distraction Filtering}
When the Sieve detects a distracted state ($S_{\text{distract}}$), the retrieved set can instead be examined to identify and remove distracting documents. Owing to its lightweight nature, the Sieve can be repeatedly applied to different document subsets; in our implementation, we perform leave-one-out filtering by removing each of the top-5 documents in turn and re-evaluating the remaining set. The cleaned document set is then forwarded to the Sage for generation.

As shown in Table~\ref{tab:post-abstention}, these strategies substantially improve accuracy on distracted examples over the no-recovery baseline.

\begin{table}[t]
\centering
\small
\begin{tabular}{llc}
\toprule
\textbf{Dataset} & \textbf{Strategy} & \textbf{Accuracy} \\
\midrule
NQ-google
& No Recovery & 0.247 \\
& \quad + Distraction Filtering & 0.867 \\
& \quad + Agent-in-the-loop & 0.716 \\
\midrule
RealtimeQA
& No Recovery & 0.380 \\
& \quad + Distraction Filtering & 0.943 \\
& \quad + Agent-in-the-loop & 0.774 \\
\bottomrule
\end{tabular}
\caption{\textbf{Accuracy on distracted examples detected by the Sieve under different post-abstention strategies.} Results are evaluated on 113 NQ-google and 71 RealtimeQA examples
identified as distracted. No Recovery denotes retaining the original abstention outcome without applying a follow-up strategy.}
\label{tab:post-abstention}
\end{table}

\section{Conclusion}
We introduced Sieve and Sage, a framework for RALM with abstention that decouples the detection of distracting document sets from core reasoning. A lightweight Longformer-based classifier (Sieve) screens retrieved document sets for distracting evidence before invoking the generative LLM (Sage), which performs grounded generation or abstains when sufficient evidence is unavailable.

Across general and expert domains, our framework achieves up to a 1.99$\times$ inference speedup and a 56.6\%p gain in system accuracy. While this leads to a slight reduction in coverage and selective accuracy, it reduces incorrect responses by up to 84.2\%, substantially improving overall reliability.

The explicit distinction between distracted and unanswerable states further supports targeted strategies such as distraction filtering and re-retrieval. 

Overall, our results demonstrate that structured control over when and why to abstain can improve the efficiency and reliability of RALMs in high-stakes expert domains.

\section*{Limitations}

\paragraph{Simulated retrieval noise}
Our evaluation primarily relies on constructed retrieval conditions, including contexts with missing evidence and simulated conflict, negated, and adversarial distractions. Although these benchmarks cover diverse forms of retrieval failure and exhibit consistent trends across datasets, they cannot fully capture the complexity of naturally occurring noise in real-world retrieval pipelines. Future work should evaluate the framework under naturally occurring retrieval failures and dynamically changing knowledge sources to establish its external validity.

\paragraph{Generalization to Zero-Shot Scenarios}
Although the Sieve adapts to expert domains with fewer than 200 labeled examples, some domain-specific supervision remains necessary to address distribution shifts in document structure and terminology. Developing a fully zero-shot, domain-agnostic Sieve that can recognize structural distractions without target-domain fine-tuning remains an important direction for future work.

\paragraph{Trade-off between abstention and selective accuracy}
The conservative response policy adopted by Sieve and Sage reduces incorrect responses but can lower coverage in some settings. This indicates a potential trade-off between avoiding harmful errors and preserving answering utility. The appropriate operating point may depend on the relative costs of incorrect answers and abstentions in the target application. Future work may explore adaptive threshold calibration and multi-objective training to better control this trade-off.

\section*{Acknowledgments}
This work was supported in part by the National Research Foundation of Korea (NRF) grant (RS-2023-00280883, RS-2023-00222663); by the National Research Foundation, Korea, under project BK21 FOUR(Dept. of Data Science, SNU, No. 5199990914569); by the Korea Institute of Science and Technology Information (KISTI) in 2026 (No. (KISTI)K26L3M1C1), aimed at developing KONI (KISTI Open Neural Intelligence), a large language model specialized in science and technology; and by the Institute of Information \& communications Technology Planning \& Evaluation (IITP) grant funded by the Korea government(MSIT) (RS-2025-02263754, Human-Centric Embodied AI Agents with Autonomous Decision-Making; RS-2025-25442149, LG AI STAR Talent Development Program for Leading Large-Scale Generative AI Models in the Physical AI Domain; and RS-2025-25463302, Global Talent Recruitment for AI-powered Drug Discovery); by grant (25202MFDS003) from Ministry of Food and Drug Safety in 2025; Institute of Information \& communications Technology Planning \& Evaluation (IITP) grant funded by the Korea government(MSIT) (No. RS-2025-25442149, LG AI STAR Talent Development Program for Leading Large-Scale Generative AI Models in the Physical AI Domain); and by the Institute of Information \& Communications Technology Planning \& Evaluation (IITP) grant funded by the Korean government (MSIT) (RS-2025-25463302, Global Talent Recruitment for AI-powered Drug Discovery).



\bibliography{latex/custom}

@misc{beltagyLongformerLongDocumentTransformer2020,
  title = {Longformer: {{The Long-Document Transformer}}},
  shorttitle = {Longformer},
  author = {Beltagy, Iz and Peters, Matthew E. and Cohan, Arman},
  year = 2020,
  month = dec,
  number = {arXiv:2004.05150},
  eprint = {2004.05150},
  primaryclass = {cs},
  publisher = {arXiv},
  doi = {10.48550/arXiv.2004.05150},
  archiveprefix = {arXiv},
}

@misc{bordesQuestionAnsweringSubgraph2014,
  title = {Question {{Answering}} with {{Subgraph Embeddings}}},
  shorttitle = {{{WebQ}}},
  author = {Bordes, Antoine and Chopra, Sumit and Weston, Jason},
  year = 2014,
  month = sep,
  number = {arXiv:1406.3676},
  eprint = {1406.3676},
  primaryclass = {cs},
  publisher = {arXiv},
  doi = {10.48550/arXiv.1406.3676},
  archiveprefix = {arXiv},
}

@misc{christopheMed42EvaluatingFineTuning2024a,
  title = {Med42 -- {{Evaluating Fine-Tuning Strategies}} for {{Medical LLMs}}: {{Full-Parameter}} vs. {{Parameter-Efficient Approaches}}},
  shorttitle = {Med42},
  author = {Christophe, Cl{\'e}ment and Kanithi, Praveen K. and Munjal, Prateek and Raha, Tathagata and Hayat, Nasir and Rajan, Ronnie and {Al-Mahrooqi}, Ahmed and Gupta, Avani and Salman, Muhammad Umar and Gosal, Gurpreet and Kanakiya, Bhargav and Chen, Charles and Vassilieva, Natalia and Amor, Boulbaba Ben and Pimentel, Marco AF and Khan, Shadab},
  year = 2024,
  month = apr,
  number = {arXiv:2404.14779},
  eprint = {2404.14779},
  primaryclass = {cs},
  publisher = {arXiv},
  doi = {10.48550/arXiv.2404.14779},
  archiveprefix = {arXiv},
}

@inproceedings{cuconasuPowerNoiseRedefining2024a,
  title = {The {{Power}} of {{Noise}}: {{Redefining Retrieval}} for {{RAG Systems}}},
  shorttitle = {The {{Power}} of {{Noise}}},
  booktitle = {Proceedings of the 47th {{International ACM SIGIR Conference}} on {{Research}} and {{Development}} in {{Information Retrieval}}},
  author = {Cuconasu, Florin and Trappolini, Giovanni and Siciliano, Federico and Filice, Simone and Campagnano, Cesare and Maarek, Yoelle and Tonellotto, Nicola and Silvestri, Fabrizio},
  year = 2024,
  month = jul,
  eprint = {2401.14887},
  primaryclass = {cs},
  pages = {719--729},
  doi = {10.1145/3626772.3657834},
  archiveprefix = {arXiv},
}

@misc{fangEnhancingNoiseRobustness2024,
  title = {Enhancing {{Noise Robustness}} of {{Retrieval-Augmented Language Models}} with {{Adaptive Adversarial Training}}},
  shorttitle = {{{RAAT}}},
  author = {Fang, Feiteng and Bai, Yuelin and Ni, Shiwen and Yang, Min and Chen, Xiaojun and Xu, Ruifeng},
  year = 2024,
  month = may,
  number = {arXiv:2405.20978},
  eprint = {2405.20978},
  primaryclass = {cs},
  publisher = {arXiv},
  doi = {10.48550/arXiv.2405.20978},
  archiveprefix = {arXiv},
  langid = {american},
}

@misc{grattafioriLlama3Herd2024a,
  title = {The {{Llama}} 3 {{Herd}} of {{Models}}},
  shorttitle = {Llama3},
  author = {Grattafiori, Aaron and Dubey, Abhimanyu and Jauhri, Abhinav and Pandey, Abhinav and Kadian, Abhishek and {Al-Dahle}, Ahmad and Letman, Aiesha and Mathur, Akhil and Schelten, Alan and Vaughan, Alex and Yang, Amy and Fan, Angela and Goyal, Anirudh and Hartshorn, Anthony and Yang, Aobo and Mitra, Archi and Sravankumar, Archie and Korenev, Artem and Hinsvark, Arthur and Rao, Arun and Zhang, Aston and Rodriguez, Aurelien and Gregerson, Austen and Spataru, Ava and Roziere, Baptiste and Biron, Bethany and Tang, Binh and Chern, Bobbie and Caucheteux, Charlotte and Nayak, Chaya and Bi, Chloe and Marra, Chris and McConnell, Chris and Keller, Christian and Touret, Christophe and Wu, Chunyang and Wong, Corinne and Ferrer, Cristian Canton and Nikolaidis, Cyrus and Allonsius, Damien and Song, Daniel and Pintz, Danielle and Livshits, Danny and Wyatt, Danny and Esiobu, David and Choudhary, Dhruv and Mahajan, Dhruv and {Garcia-Olano}, Diego and Perino, Diego and Hupkes, Dieuwke and Lakomkin, Egor and AlBadawy, Ehab and Lobanova, Elina and Dinan, Emily and Smith, Eric Michael and Radenovic, Filip and Guzm{\'a}n, Francisco and Zhang, Frank and Synnaeve, Gabriel and Lee, Gabrielle and Anderson, Georgia Lewis and Thattai, Govind and Nail, Graeme and Mialon, Gregoire and Pang, Guan and Cucurell, Guillem and Nguyen, Hailey and Korevaar, Hannah and Xu, Hu and Touvron, Hugo and Zarov, Iliyan and Ibarra, Imanol Arrieta and Kloumann, Isabel and Misra, Ishan and Evtimov, Ivan and Zhang, Jack and Copet, Jade and Lee, Jaewon and Geffert, Jan and Vranes, Jana and Park, Jason and Mahadeokar, Jay and Shah, Jeet and van der Linde, Jelmer and Billock, Jennifer and Hong, Jenny and Lee, Jenya and Fu, Jeremy and Chi, Jianfeng and Huang, Jianyu and Liu, Jiawen and Wang, Jie and Yu, Jiecao and Bitton, Joanna and Spisak, Joe and Park, Jongsoo and Rocca, Joseph and Johnstun, Joshua and Saxe, Joshua and Jia, Junteng and Alwala, Kalyan Vasuden and Prasad, Karthik and Upasani, Kartikeya and Plawiak, Kate and Li, Ke and Heafield, Kenneth and Stone, Kevin and {El-Arini}, Khalid and Iyer, Krithika and Malik, Kshitiz and Chiu, Kuenley and Bhalla, Kunal and Lakhotia, Kushal and {Rantala-Yeary}, Lauren and van der Maaten, Laurens and Chen, Lawrence and Tan, Liang and Jenkins, Liz and Martin, Louis and Madaan, Lovish and Malo, Lubo and Blecher, Lukas and Landzaat, Lukas and de Oliveira, Luke and Muzzi, Madeline and Pasupuleti, Mahesh and Singh, Mannat and Paluri, Manohar and Kardas, Marcin and Tsimpoukelli, Maria and Oldham, Mathew and Rita, Mathieu and Pavlova, Maya and Kambadur, Melanie and Lewis, Mike and Si, Min and Singh, Mitesh Kumar and Hassan, Mona and Goyal, Naman and Torabi, Narjes and Bashlykov, Nikolay and Bogoychev, Nikolay and Chatterji, Niladri and Zhang, Ning and Duchenne, Olivier and {\c C}elebi, Onur and Alrassy, Patrick and Zhang, Pengchuan and Li, Pengwei and Vasic, Petar and Weng, Peter and Bhargava, Prajjwal and Dubal, Pratik and Krishnan, Praveen and Koura, Punit Singh and Xu, Puxin and He, Qing and Dong, Qingxiao and Srinivasan, Ragavan and Ganapathy, Raj and Calderer, Ramon and Cabral, Ricardo Silveira and Stojnic, Robert and Raileanu, Roberta and Maheswari, Rohan and Girdhar, Rohit and Patel, Rohit and Sauvestre, Romain and Polidoro, Ronnie and Sumbaly, Roshan and Taylor, Ross and Silva, Ruan and Hou, Rui and Wang, Rui and Hosseini, Saghar and Chennabasappa, Sahana and Singh, Sanjay and Bell, Sean and Kim, Seohyun Sonia and Edunov, Sergey and Nie, Shaoliang and Narang, Sharan and Raparthy, Sharath and Shen, Sheng and Wan, Shengye and Bhosale, Shruti and Zhang, Shun and Vandenhende, Simon and Batra, Soumya and Whitman, Spencer and Sootla, Sten and Collot, Stephane and Gururangan, Suchin and Borodinsky, Sydney and Herman, Tamar and Fowler, Tara and Sheasha, Tarek and Georgiou, Thomas and Scialom, Thomas and Speckbacher, Tobias and Mihaylov, Todor and Xiao, Tong and Karn, Ujjwal and Goswami, Vedanuj and Gupta, Vibhor and Ramanathan, Vignesh and Kerkez, Viktor and Gonguet, Vincent and Do, Virginie and Vogeti, Vish and Albiero, V{\'i}tor and Petrovic, Vladan and Chu, Weiwei and Xiong, Wenhan and Fu, Wenyin and Meers, Whitney and Martinet, Xavier and Wang, Xiaodong and Wang, Xiaofang and Tan, Xiaoqing Ellen and Xia, Xide and Xie, Xinfeng and Jia, Xuchao and Wang, Xuewei and Goldschlag, Yaelle and Gaur, Yashesh and Babaei, Yasmine and Wen, Yi and Song, Yiwen and Zhang, Yuchen and Li, Yue and Mao, Yuning and Coudert, Zacharie Delpierre and Yan, Zheng and Chen, Zhengxing and Papakipos, Zoe and Singh, Aaditya and Srivastava, Aayushi and Jain, Abha and Kelsey, Adam and Shajnfeld, Adam and Gangidi, Adithya and Victoria, Adolfo and Goldstand, Ahuva and Menon, Ajay and Sharma, Ajay and Boesenberg, Alex and Baevski, Alexei and Feinstein, Allie and Kallet, Amanda and Sangani, Amit and Teo, Amos and Yunus, Anam and Lupu, Andrei and Alvarado, Andres and Caples, Andrew and Gu, Andrew and Ho, Andrew and Poulton, Andrew and Ryan, Andrew and Ramchandani, Ankit and Dong, Annie and Franco, Annie and Goyal, Anuj and Saraf, Aparajita and Chowdhury, Arkabandhu and Gabriel, Ashley and Bharambe, Ashwin and Eisenman, Assaf and Yazdan, Azadeh and James, Beau and Maurer, Ben and Leonhardi, Benjamin and Huang, Bernie and Loyd, Beth and Paola, Beto De and Paranjape, Bhargavi and Liu, Bing and Wu, Bo and Ni, Boyu and Hancock, Braden and Wasti, Bram and Spence, Brandon and Stojkovic, Brani and Gamido, Brian and Montalvo, Britt and Parker, Carl and Burton, Carly and Mejia, Catalina and Liu, Ce and Wang, Changhan and Kim, Changkyu and Zhou, Chao and Hu, Chester and Chu, Ching-Hsiang and Cai, Chris and Tindal, Chris and Feichtenhofer, Christoph and Gao, Cynthia and Civin, Damon and Beaty, Dana and Kreymer, Daniel and Li, Daniel and Adkins, David and Xu, David and Testuggine, Davide and David, Delia and Parikh, Devi and Liskovich, Diana and Foss, Didem and Wang, Dingkang and Le, Duc and Holland, Dustin and Dowling, Edward and Jamil, Eissa and Montgomery, Elaine and Presani, Eleonora and Hahn, Emily and Wood, Emily and Le, Eric-Tuan and Brinkman, Erik and Arcaute, Esteban and Dunbar, Evan and Smothers, Evan and Sun, Fei and Kreuk, Felix and Tian, Feng and Kokkinos, Filippos and Ozgenel, Firat and Caggioni, Francesco and Kanayet, Frank and Seide, Frank and Florez, Gabriela Medina and Schwarz, Gabriella and Badeer, Gada and Swee, Georgia and Halpern, Gil and Herman, Grant and Sizov, Grigory and Guangyi and Zhang and Lakshminarayanan, Guna and Inan, Hakan and Shojanazeri, Hamid and Zou, Han and Wang, Hannah and Zha, Hanwen and Habeeb, Haroun and Rudolph, Harrison and Suk, Helen and Aspegren, Henry and Goldman, Hunter and Zhan, Hongyuan and Damlaj, Ibrahim and Molybog, Igor and Tufanov, Igor and Leontiadis, Ilias and Veliche, Irina-Elena and Gat, Itai and Weissman, Jake and Geboski, James and Kohli, James and Lam, Janice and Asher, Japhet and Gaya, Jean-Baptiste and Marcus, Jeff and Tang, Jeff and Chan, Jennifer and Zhen, Jenny and Reizenstein, Jeremy and Teboul, Jeremy and Zhong, Jessica and Jin, Jian and Yang, Jingyi and Cummings, Joe and Carvill, Jon and Shepard, Jon and McPhie, Jonathan and Torres, Jonathan and Ginsburg, Josh and Wang, Junjie and Wu, Kai and U, Kam Hou and Saxena, Karan and Khandelwal, Kartikay and Zand, Katayoun and Matosich, Kathy and Veeraraghavan, Kaushik and Michelena, Kelly and Li, Keqian and Jagadeesh, Kiran and Huang, Kun and Chawla, Kunal and Huang, Kyle and Chen, Lailin and Garg, Lakshya and A, Lavender and Silva, Leandro and Bell, Lee and Zhang, Lei and Guo, Liangpeng and Yu, Licheng and Moshkovich, Liron and Wehrstedt, Luca and Khabsa, Madian and Avalani, Manav and Bhatt, Manish and Mankus, Martynas and Hasson, Matan and Lennie, Matthew and Reso, Matthias and Groshev, Maxim and Naumov, Maxim and Lathi, Maya and Keneally, Meghan and Liu, Miao and Seltzer, Michael L. and Valko, Michal and Restrepo, Michelle and Patel, Mihir and Vyatskov, Mik and Samvelyan, Mikayel and Clark, Mike and Macey, Mike and Wang, Mike and Hermoso, Miquel Jubert and Metanat, Mo and Rastegari, Mohammad and Bansal, Munish and Santhanam, Nandhini and Parks, Natascha and White, Natasha and Bawa, Navyata and Singhal, Nayan and Egebo, Nick and Usunier, Nicolas and Mehta, Nikhil and Laptev, Nikolay Pavlovich and Dong, Ning and Cheng, Norman and Chernoguz, Oleg and Hart, Olivia and Salpekar, Omkar and Kalinli, Ozlem and Kent, Parkin and Parekh, Parth and Saab, Paul and Balaji, Pavan and Rittner, Pedro and Bontrager, Philip and Roux, Pierre and Dollar, Piotr and Zvyagina, Polina and Ratanchandani, Prashant and Yuvraj, Pritish and Liang, Qian and Alao, Rachad and Rodriguez, Rachel and Ayub, Rafi and Murthy, Raghotham and Nayani, Raghu and Mitra, Rahul and Parthasarathy, Rangaprabhu and Li, Raymond and Hogan, Rebekkah and Battey, Robin and Wang, Rocky and Howes, Russ and Rinott, Ruty and Mehta, Sachin and Siby, Sachin and Bondu, Sai Jayesh and Datta, Samyak and Chugh, Sara and Hunt, Sara and Dhillon, Sargun and Sidorov, Sasha and Pan, Satadru and Mahajan, Saurabh and Verma, Saurabh and Yamamoto, Seiji and Ramaswamy, Sharadh and Lindsay, Shaun and Lindsay, Shaun and Feng, Sheng and Lin, Shenghao and Zha, Shengxin Cindy and Patil, Shishir and Shankar, Shiva and Zhang, Shuqiang and Zhang, Shuqiang and Wang, Sinong and Agarwal, Sneha and Sajuyigbe, Soji and Chintala, Soumith and Max, Stephanie and Chen, Stephen and Kehoe, Steve and Satterfield, Steve and Govindaprasad, Sudarshan and Gupta, Sumit and Deng, Summer and Cho, Sungmin and Virk, Sunny and Subramanian, Suraj and Choudhury, Sy and Goldman, Sydney and Remez, Tal and Glaser, Tamar and Best, Tamara and Koehler, Thilo and Robinson, Thomas and Li, Tianhe and Zhang, Tianjun and Matthews, Tim and Chou, Timothy and Shaked, Tzook and Vontimitta, Varun and Ajayi, Victoria and Montanez, Victoria and Mohan, Vijai and Kumar, Vinay Satish and Mangla, Vishal and Ionescu, Vlad and Poenaru, Vlad and Mihailescu, Vlad Tiberiu and Ivanov, Vladimir and Li, Wei and Wang, Wenchen and Jiang, Wenwen and Bouaziz, Wes and Constable, Will and Tang, Xiaocheng and Wu, Xiaojian and Wang, Xiaolan and Wu, Xilun and Gao, Xinbo and Kleinman, Yaniv and Chen, Yanjun and Hu, Ye and Jia, Ye and Qi, Ye and Li, Yenda and Zhang, Yilin and Zhang, Ying and Adi, Yossi and Nam, Youngjin and Yu and Wang and Zhao, Yu and Hao, Yuchen and Qian, Yundi and Li, Yunlu and He, Yuzi and Rait, Zach and DeVito, Zachary and Rosnbrick, Zef and Wen, Zhaoduo and Yang, Zhenyu and Zhao, Zhiwei and Ma, Zhiyu},
  year = 2024,
  month = nov,
  number = {arXiv:2407.21783},
  eprint = {2407.21783},
  primaryclass = {cs},
  publisher = {arXiv},
  doi = {10.48550/arXiv.2407.21783},
  archiveprefix = {arXiv},
}

@misc{gul2026mashmodelingabstentionselective,
      title={MASH: Modeling Abstention via Selective Help-Seeking}, 
      author={Mustafa Omer Gul and Claire Cardie and Tanya Goyal},
      year={2026},
      eprint={2510.01152},
      archivePrefix={arXiv},
      primaryClass={cs.CL},
      url={https://arxiv.org/abs/2510.01152}, 
}

@article{haanAstroMLab3Achieving2025,
  title = {{{AstroMLab}} 3: {{Achieving GPT-4o Level Performance}} in {{Astronomy}} with a {{Specialized 8B-Parameter Large Language Model}}},
  shorttitle = {{{AstroMLab}} 3},
  author = {de Haan, Tijmen and Ting, Yuan-Sen and Ghosal, Tirthankar and Nguyen, Tuan Dung and Accomazzi, Alberto and Wells, Azton and Ramachandra, Nesar and Pan, Rui and Sun, Zechang},
  year = 2025,
  month = apr,
  journal = {Scientific Reports},
  volume = {15},
  number = {1},
  eprint = {2411.09012},
  primaryclass = {astro-ph},
  pages = {13751},
  issn = {2045-2322},
  doi = {10.1038/s41598-025-97131-y},
  archiveprefix = {arXiv},
}

@misc{jorenSufficientContextNew2025a,
  title = {Sufficient {{Context}}: {{A New Lens}} on {{Retrieval Augmented Generation Systems}}},
  shorttitle = {Sufficient {{Context}}},
  author = {Joren, Hailey and Zhang, Jianyi and Ferng, Chun-Sung and Juan, Da-Cheng and Taly, Ankur and Rashtchian, Cyrus},
  year = 2025,
  month = apr,
  number = {arXiv:2411.06037},
  eprint = {2411.06037},
  primaryclass = {cs},
  publisher = {arXiv},
  doi = {10.48550/arXiv.2411.06037},
  archiveprefix = {arXiv},
}

@misc{kimRERAGImprovingOpenDomain2024a,
  title = {{{RE-RAG}}: {{Improving Open-Domain QA Performance}} and {{Interpretability}} with {{Relevance Estimator}} in {{Retrieval-Augmented Generation}}},
  shorttitle = {{{RE-RAG}}},
  author = {Kim, Kiseung and Lee, Jay-Yoon},
  year = 2024,
  month = oct,
  number = {arXiv:2406.05794},
  eprint = {2406.05794},
  primaryclass = {cs},
  publisher = {arXiv},
  doi = {10.48550/arXiv.2406.05794},
  archiveprefix = {arXiv},
}

@inproceedings{leeLatentRetrievalWeakly2019,
  title = {Latent {{Retrieval}} for {{Weakly Supervised Open Domain Question Answering}}},
  shorttitle = {{{NQ-open}}},
  booktitle = {Proceedings of the 57th {{Annual Meeting}} of the {{Association}} for {{Computational Linguistics}}},
  author = {Lee, Kenton and Chang, Ming-Wei and Toutanova, Kristina},
  editor = {Korhonen, Anna and Traum, David and M{\`a}rquez, Llu{\'i}s},
  year = 2019,
  month = jul,
  pages = {6086--6096},
  publisher = {Association for Computational Linguistics},
  address = {Florence, Italy},
  doi = {10.18653/v1/P19-1612},
  langid = {american},
}

@inproceedings{liThinkTwiceTrusting2024,
  title = {Think {{Twice Before Trusting}}: {{Self-Detection}} for {{Large Language Models}} through {{Comprehensive Answer Reflection}}},
  shorttitle = {Think {{Twice Before Trusting}}},
  booktitle = {Findings of the {{Association}} for {{Computational Linguistics}}: {{EMNLP}} 2024},
  author = {Li, Moxin and Wang, Wenjie and Feng, Fuli and Zhu, Fengbin and Wang, Qifan and Chua, Tat-Seng},
  editor = {{Al-Onaizan}, Yaser and Bansal, Mohit and Chen, Yun-Nung},
  year = 2024,
  month = jan,
  pages = {11858--11875},
  publisher = {Association for Computational Linguistics},
  address = {Miami, Florida, USA},
  doi = {10.18653/v1/2024.findings-emnlp.693},
}

@misc{palMedMCQALargescaleMultiSubject2022,
  title = {{{MedMCQA}} : {{A Large-scale Multi-Subject Multi-Choice Dataset}} for {{Medical}} Domain {{Question Answering}}},
  shorttitle = {{{MedMCQA}}},
  author = {Pal, Ankit and Umapathi, Logesh Kumar and Sankarasubbu, Malaikannan},
  year = 2022,
  month = mar,
  number = {arXiv:2203.14371},
  eprint = {2203.14371},
  primaryclass = {cs},
  publisher = {arXiv},
  doi = {10.48550/arXiv.2203.14371},
  archiveprefix = {arXiv},
}

@inproceedings{parkEnhancingRobustnessRetrievalAugmented,
    title = "Enhancing Robustness of Retrieval-Augmented Language Models with In-Context Learning",
    author = "Park, SeongIl  and
      Choi, Seungwoo  and
      Kim, Nahyun  and
      Lee, Jay-Yoon",
    editor = "Yu, Wenhao  and
      Shi, Weijia  and
      Yasunaga, Michihiro  and
      Jiang, Meng  and
      Zhu, Chenguang  and
      Hajishirzi, Hannaneh  and
      Zettlemoyer, Luke  and
      Zhang, Zhihan",
    booktitle = "Proceedings of the 3rd Workshop on Knowledge Augmented Methods for NLP",
    month = aug,
    year = "2024",
    address = "Bangkok, Thailand",
    publisher = "Association for Computational Linguistics",
    url = "https://aclanthology.org/2024.knowledgenlp-1.7/",
    doi = "10.18653/v1/2024.knowledgenlp-1.7",
    pages = "93--102",
}

@misc{ramInContextRetrievalAugmentedLanguage2023a,
  title = {In-{{Context Retrieval-Augmented Language Models}}},
  shorttitle = {In-Context {{RALM}}},
  author = {Ram, Ori and Levine, Yoav and Dalmedigos, Itay and Muhlgay, Dor and Shashua, Amnon and {Leyton-Brown}, Kevin and Shoham, Yoav},
  year = 2023,
  month = aug,
  number = {arXiv:2302.00083},
  eprint = {2302.00083},
  primaryclass = {cs},
  publisher = {arXiv},
  doi = {10.48550/arXiv.2302.00083},
  archiveprefix = {arXiv},
  langid = {american},
}

@misc{santhanamColBERTv2EffectiveEfficient2022a,
  title = {{{ColBERTv2}}: {{Effective}} and {{Efficient Retrieval}} via {{Lightweight Late Interaction}}},
  shorttitle = {{{ColBERTv2}}},
  author = {Santhanam, Keshav and Khattab, Omar and {Saad-Falcon}, Jon and Potts, Christopher and Zaharia, Matei},
  year = 2022,
  month = jul,
  number = {arXiv:2112.01488},
  eprint = {2112.01488},
  primaryclass = {cs},
  publisher = {arXiv},
  doi = {10.48550/arXiv.2112.01488},
  archiveprefix = {arXiv},
}

@inproceedings{shiREPLUGRetrievalAugmentedBlackBox2024,
  title = {{{REPLUG}}: {{Retrieval-Augmented Black-Box Language Models}}},
  shorttitle = {{{REPLUG}}},
  booktitle = {Proceedings of the 2024 {{Conference}} of the {{North American Chapter}} of the {{Association}} for {{Computational Linguistics}}: {{Human Language Technologies}} ({{Volume}} 1: {{Long Papers}})},
  author = {Shi, Weijia and Min, Sewon and Yasunaga, Michihiro and Seo, Minjoon and James, Richard and Lewis, Mike and Zettlemoyer, Luke and Yih, Wen-tau},
  editor = {Duh, Kevin and Gomez, Helena and Bethard, Steven},
  year = 2024,
  month = jun,
  pages = {8371--8384},
  publisher = {Association for Computational Linguistics},
  address = {Mexico City, Mexico},
  doi = {10.18653/v1/2024.naacl-long.463},
}

@inproceedings{sunDivideThenAlignHonestAlignment2025,
  title = {Divide-{{Then-Align}}: {{Honest Alignment}} Based on the {{Knowledge Boundary}} of {{RAG}}},
  shorttitle = {Divide-{{Then-Align}}},
  booktitle = {Proceedings of the 63rd {{Annual Meeting}} of the {{Association}} for {{Computational Linguistics}} ({{Volume}} 1: {{Long Papers}})},
  author = {Sun, Xin and Xie, Jianan and Chen, Zhongqi and Liu, Qiang and Wu, Shu and Chen, Yuehe and Song, Bowen and Wang, Zilei and Wang, Weiqiang and Wang, Liang},
  editor = {Che, Wanxiang and Nabende, Joyce and Shutova, Ekaterina and Pilehvar, Mohammad Taher},
  year = 2025,
  month = jul,
  pages = {11461--11480},
  publisher = {Association for Computational Linguistics},
  address = {Vienna, Austria},
  doi = {10.18653/v1/2025.acl-long.561},
  isbn = {979-8-89176-251-0},
}

@inproceedings{thakurKnowingWhenYou2024,
  title = {``{{Knowing When You Don}}'t {{Know}}'': {{A Multilingual Relevance Assessment Dataset}} for {{Robust Retrieval-Augmented Generation}}},
  shorttitle = {{{NoMIRACL}}},
  booktitle = {Findings of the {{Association}} for {{Computational Linguistics}}: {{EMNLP}} 2024},
  author = {Thakur, Nandan and Bonifacio, Luiz and Zhang, Crystina and Ogundepo, Odunayo and Kamalloo, Ehsan and {Alfonso-Hermelo}, David and Li, Xiaoguang and Liu, Qun and Chen, Boxing and Rezagholizadeh, Mehdi and Lin, Jimmy},
  editor = {{Al-Onaizan}, Yaser and Bansal, Mohit and Chen, Yun-Nung},
  year = 2024,
  month = jan,
  pages = {12508--12526},
  publisher = {Association for Computational Linguistics},
  address = {Miami, Florida, USA},
  doi = {10.18653/v1/2024.findings-emnlp.730},
}

@misc{wangDomainRAGChineseBenchmark2024b,
  title = {{{DomainRAG}}: {{A Chinese Benchmark}} for {{Evaluating Domain-specific Retrieval-Augmented Generation}}},
  shorttitle = {{{DomainRAG}}},
  author = {Wang, Shuting and Liu, Jiongnan and Song, Shiren and Cheng, Jiehan and Fu, Yuqi and Guo, Peidong and Fang, Kun and Zhu, Yutao and Dou, Zhicheng},
  year = 2024,
  month = jun,
  number = {arXiv:2406.05654},
  eprint = {2406.05654},
  primaryclass = {cs},
  publisher = {arXiv},
  doi = {10.48550/arXiv.2406.05654},
  archiveprefix = {arXiv},
}

@misc{xiangCertifiablyRobustRAG2024a,
  title = {Certifiably {{Robust RAG}} against {{Retrieval Corruption}}},
  shorttitle = {{{CRRAG}}},
  author = {Xiang, Chong and Wu, Tong and Zhong, Zexuan and Wagner, David and Chen, Danqi and Mittal, Prateek},
  year = 2024,
  month = may,
  number = {arXiv:2405.15556},
  eprint = {2405.15556},
  primaryclass = {cs},
  publisher = {arXiv},
  doi = {10.48550/arXiv.2405.15556},
  archiveprefix = {arXiv},
}

@misc{yanCorrectiveRetrievalAugmented2024,
  title = {Corrective {{Retrieval Augmented Generation}}},
  shorttitle = {{{CRAG}}},
  author = {Yan, Shi-Qi and Gu, Jia-Chen and Zhu, Yun and Ling, Zhen-Hua},
  year = 2024,
  month = oct,
  number = {arXiv:2401.15884},
  eprint = {2401.15884},
  primaryclass = {cs},
  publisher = {arXiv},
  doi = {10.48550/arXiv.2401.15884},
  archiveprefix = {arXiv},
  langid = {american},
}

@misc{yangHotpotQADatasetDiverse2018,
  title = {{{HotpotQA}}: {{A Dataset}} for {{Diverse}}, {{Explainable Multi-hop Question Answering}}},
  shorttitle = {{{HotpotQA}}},
  author = {Yang, Zhilin and Qi, Peng and Zhang, Saizheng and Bengio, Yoshua and Cohen, William W. and Salakhutdinov, Ruslan and Manning, Christopher D.},
  year = 2018,
  month = sep,
  number = {arXiv:1809.09600},
  eprint = {1809.09600},
  primaryclass = {cs},
  publisher = {arXiv},
  doi = {10.48550/arXiv.1809.09600},
  archiveprefix = {arXiv},
}

@misc{yoranMakingRetrievalAugmentedLanguage2024,
  title = {Making {{Retrieval-Augmented Language Models Robust}} to {{Irrelevant Context}}},
  shorttitle = {{{RetRobust}}},
  author = {Yoran, Ori and Wolfson, Tomer and Ram, Ori and Berant, Jonathan},
  year = 2024,
  month = may,
  number = {arXiv:2310.01558},
  eprint = {2310.01558},
  primaryclass = {cs},
  publisher = {arXiv},
  doi = {10.48550/arXiv.2310.01558},
  archiveprefix = {arXiv},
  langid = {american},
}

@misc{yuChainofNoteEnhancingRobustness2024,
  title = {Chain-of-{{Note}}: {{Enhancing Robustness}} in {{Retrieval-Augmented Language Models}}},
  shorttitle = {Chain-of-{{Note}}},
  author = {Yu, Wenhao and Zhang, Hongming and Pan, Xiaoman and Ma, Kaixin and Wang, Hongwei and Yu, Dong},
  year = 2024,
  month = oct,
  number = {arXiv:2311.09210},
  eprint = {2311.09210},
  primaryclass = {cs},
  publisher = {arXiv},
  doi = {10.48550/arXiv.2311.09210},
  archiveprefix = {arXiv},
}

@misc{zhangRAFTAdaptingLanguage2024,
  title = {{{RAFT}}: {{Adapting Language Model}} to {{Domain Specific RAG}}},
  shorttitle = {{{RAFT}}},
  author = {Zhang, Tianjun and Patil, Shishir G. and Jain, Naman and Shen, Sheng and Zaharia, Matei and Stoica, Ion and Gonzalez, Joseph E.},
  year = 2024,
  month = jun,
  number = {arXiv:2403.10131},
  eprint = {2403.10131},
  primaryclass = {cs},
  publisher = {arXiv},
  doi = {10.48550/arXiv.2403.10131},
  archiveprefix = {arXiv},
}

@inproceedings{zhongPoisoningRetrievalCorpora2023,
  title = {Poisoning {{Retrieval Corpora}} by {{Injecting Adversarial Passages}}},
  shorttitle = {Poisoned {{Corpus}}},
  booktitle = {Proceedings of the 2023 {{Conference}} on {{Empirical Methods}} in {{Natural Language Processing}}},
  author = {Zhong, Zexuan and Huang, Ziqing and Wettig, Alexander and Chen, Danqi},
  editor = {Bouamor, Houda and Pino, Juan and Bali, Kalika},
  year = 2023,
  month = feb,
  pages = {13764--13775},
  publisher = {Association for Computational Linguistics},
  address = {Singapore},
  doi = {10.18653/v1/2023.emnlp-main.849},
  langid = {american},
}

@misc{zouPoisonedRAGKnowledgeCorruption2024,
  title = {{{PoisonedRAG}}: {{Knowledge Corruption Attacks}} to {{Retrieval-Augmented Generation}} of {{Large Language Models}}},
  shorttitle = {{{PoisonedRAG}}},
  author = {Zou, Wei and Geng, Runpeng and Wang, Binghui and Jia, Jinyuan},
  year = 2024,
  month = aug,
  number = {arXiv:2402.07867},
  eprint = {2402.07867},
  primaryclass = {cs},
  publisher = {arXiv},
  doi = {10.48550/arXiv.2402.07867},
  archiveprefix = {arXiv},
  langid = {american},
}

@inproceedings{hong-etal-2024-gullible,
    title = "Why So Gullible? Enhancing the Robustness of Retrieval-Augmented Models against Counterfactual Noise",
    author = "Hong, Giwon  and
      Kim, Jeonghwan  and
      Kang, Junmo  and
      Myaeng, Sung-Hyon  and
      Whang, Joyce Jiyoung",
    editor = "Duh, Kevin  and
      Gomez, Helena  and
      Bethard, Steven",
    booktitle = "Findings of the Association for Computational Linguistics: NAACL 2024",
    month = jun,
    year = "2024",
    address = "Mexico City, Mexico",
    publisher = "Association for Computational Linguistics",
    url = "https://aclanthology.org/2024.findings-naacl.159/",
    doi = "10.18653/v1/2024.findings-naacl.159",
    pages = "2474--2495",
}

@misc{hu2021loralowrankadaptationlarge,
      title={LoRA: Low-Rank Adaptation of Large Language Models}, 
      author={Edward J. Hu and Yelong Shen and Phillip Wallis and Zeyuan Allen-Zhu and Yuanzhi Li and Shean Wang and Lu Wang and Weizhu Chen},
      year={2021},
      eprint={2106.09685},
      archivePrefix={arXiv},
      primaryClass={cs.CL},
      url={https://arxiv.org/abs/2106.09685}, 
}

@misc{nentidis2025overviewbioasq2025thirteenth,
      title={Overview of BioASQ 2025: The Thirteenth BioASQ Challenge on Large-Scale Biomedical Semantic Indexing and Question Answering}, 
      author={Anastasios Nentidis and Georgios Katsimpras and Anastasia Krithara and Martin Krallinger and Miguel Rodríguez-Ortega and Eduard Rodriguez-López and Natalia Loukachevitch and Andrey Sakhovskiy and Elena Tutubalina and Dimitris Dimitriadis and Grigorios Tsoumakas and George Giannakoulas and Alexandra Bekiaridou and Athanasios Samaras and Giorgio Maria Di Nunzio and Nicola Ferro and Stefano Marchesin and Marco Martinelli and Gianmaria Silvello and Georgios Paliouras},
      year={2025},
      eprint={2508.20554},
      archivePrefix={arXiv},
      primaryClass={cs.CL},
      url={https://arxiv.org/abs/2508.20554}, 
}

\appendix

\section{Validation of Simulated Distracting Documents} \label{sec:validation-of-simulated-distracting-documents}

\begin{table*}[t!]
\centering
\begin{subtable}[t]{\textwidth}

\resizebox{\textwidth}{!}{%
\begin{tabular}{@{}lccccccc@{}}
\toprule
 & \multicolumn{4}{c}{\textbf{General domain}} & \multicolumn{3}{c}{\textbf{Expert domain}} \\ 
\cmidrule(lr){2-5} \cmidrule(lr){6-8}
Datasets           & NQ-google & NQ-wiki & WebQ & RealtimeQA & MedQA & AstroQA & BioASQ \\
\midrule
\# Passage          & 10 $\sim$33 & 8 & 8 & 27 $\sim$53 & 24 & 24 & 4 $\sim$36 \\ \midrule
Gold Passage        & - & - & - & - & 3.714 & 6.350 & - \\
Conflict Passage    & 4.560 & 3.028 & 3.364 & 2.970 & 2.227 & 3.560 & 7.699 \\
Adversarial Passage & 19.520 & 5.700 & 6.654 & 26.490 & 3.548 & 9.610 & 7.758 \\
Negated Passage     & 5.400 & 2.659 & 2.705 & 17.160 & 1.813 & 2.710 & 2.415 \\
\bottomrule
\end{tabular}%
}
\caption{Mean Raw Rank}
\vspace{0.2cm}
\label{tab:rerank_raw}
\end{subtable}
\begin{subtable}[t]{\textwidth}
\centering
\resizebox{\textwidth}{!}{%
\begin{tabular}{@{}lccccccc@{}}
\toprule
 & \multicolumn{4}{c}{\textbf{General domain}} & \multicolumn{3}{c}{\textbf{Expert domain}} \\ 
\cmidrule(lr){2-5} \cmidrule(lr){6-8}
Datasets           & NQ-google & NQ-wiki & WebQ & RealtimeQA & MedQA & AstroQA & BioASQ \\
\midrule
\# Passage          & 10 $\sim$33 & 8 & 8 & 27 $\sim$53 & 24 & 24 & 4 $\sim$36 \\ \midrule
Gold Passage        & - & - & - & - & 0.118 & 0.233 & - \\
Conflict Passage    & 0.121 & 0.290 & 0.338 & 0.041 & 0.101 & 0.111 & 0.665 \\
Adversarial Passage & 0.598 & 0.709 & 0.854 & 0.533 & 0.318 & 0.374 & 0.643 \\
Negated Passage     & 0.145 & 0.248 & 0.258 & 0.335 & 0.080 & 0.075 & 0.139 \\
\bottomrule
\end{tabular}%
}
\caption{Mean Scaled Rank}
\label{tab:rerank_scaled}
\end{subtable}
\caption{\textbf{Ranking results of simulated distracting documents}. We evaluate the visibility of simulated noise---Conflict, Adversarial, and Negated passages---by mixing them with originally retrieved documents and reranking the set. The high rankings of these distractions demonstrate that our simulated noise is semantically plausible and poses a realistic challenge to RALMs.}
\label{tab:rerank}

\end{table*}

\begin{table*}[t]
\centering
\resizebox{\textwidth}{!}{%
\begin{tabular}{@{}lrrrrrr@{}}
\toprule
dataset &
  \multicolumn{1}{l}{\# examples} &
  \multicolumn{1}{l}{Acc (ans)} &
  \multicolumn{1}{l}{Acc (unans)} &
  \multicolumn{1}{l}{Acc (conflict)} &
  \multicolumn{1}{l}{Acc (negated)} &
  \multicolumn{1}{l}{Acc (adversarial)} \\ \midrule
NQ-google  & 30  & 23.33 & 6.67~\textcolor{red}{\scriptsize -16.66} & 20.00~\textcolor{red}{\scriptsize -3.33} & 43.33~\textcolor{green!70!black}{\scriptsize +20.00} & 13.33~\textcolor{red}{\scriptsize -10.00} \\
NQ-wiki    & 327 & 32.42 & 0.00~\textcolor{red}{\scriptsize -32.42} & 24.46~\textcolor{red}{\scriptsize -7.96} & 54.74~\textcolor{green!70!black}{\scriptsize +22.32} & 29.97~\textcolor{red}{\scriptsize -2.45} \\
WebQ       & 264 & 24.24 & 0.76~\textcolor{red}{\scriptsize -23.48} & 20.45~\textcolor{red}{\scriptsize -3.79} & 46.21~\textcolor{green!70!black}{\scriptsize +21.97} & 22.35~\textcolor{red}{\scriptsize -1.89} \\
RealtimeQA & 25  & 24.00 & 4.00~\textcolor{red}{\scriptsize -20.00} & 8.00~\textcolor{red}{\scriptsize -16.00} & 48.00~\textcolor{green!70!black}{\scriptsize +24.00} & 16.00~\textcolor{red}{\scriptsize -8.00} \\
MedQA      & 57  & 35.09 & 14.04~\textcolor{red}{\scriptsize -21.05} & 22.81~\textcolor{red}{\scriptsize -12.28} & 49.12~\textcolor{green!70!black}{\scriptsize +14.03} & 24.56~\textcolor{red}{\scriptsize -10.53} \\ \bottomrule
\end{tabular}%
}
\caption{\textbf{QA accuracy of a vanilla LLM under different retrieval conditions}. Each column reports accuracy with a specific type of retrieval state, all derived from the same answerable retrieval context by modifying one document.}
\label{tab:apple-to-apple}
\end{table*}

\subsection{Ranking Distribution of Simulated Distractions}
To ensure that our simulated distracting documents---Conflict, Adversarial, and Negated---pose a realistic challenge to RALMs, we inserted these simulated passages into the original pool of 20 retrieved documents and performed reranking using \texttt{BAAI/bge-reranker-large}\footnote{\href{https://huggingface.co/BAAI/bge-reranker-large}{https://huggingface.co/BAAI/bge-reranker-large}}. If these simulated passages are ranked high, it suggests that they are semantically similar to naturally retrieved documents and can therefore function as realistic distractions.

As summarized in Table~\ref{tab:rerank}, the simulated distractors are consistently ranked high across both general and expert-domain datasets. In particular, Negated and Conflict documents frequently appear near the top of the rankings, often within the top 10\%–30\% of the retrieved set (e.g., Mean Scaled Rank values of 0.080 and 0.101 in MedQA, respectively). Although Adversarial passages sometimes rank slightly lower (e.g., 0.854 in WebQ) because the intentionally misleading entities slightly alter their dense representations, they still remain an important target of our framework.

Overall, all types of distractions appear alongside genuine passages in both general and expert domains. This result shows that standard semantic retrievers have difficulty distinguishing logical inconsistencies from superficial semantic similarity, highlighting the need for our logic-based Sieve module for post-retrieval filtering.

\subsection{Probing Distraction Complexity: From Pairwise Clarity to Multi-document Interference}
To validate the realism and logical integrity of our simulated distracting documents, we conducted a two-stage probing experiment. 

\paragraph{Pairwise Logical Verification}
First, we presented a state-of-the-art LLM (GPT-4o) with a pair of documents: the gold document and one additional document, which could be either a simulated distractor (e.g., negated, conflicting, or adversarial) or a neutral document. When asked to categorize the relationship between the two documents, the model achieved a near 100\% accuracy. 

\paragraph{Multi-document Distraction Effect} 
However, Table~\ref{tab:full_results-models} shows that the classification accuracy dropped significantly when we increased the complexity by providing the full set of five documents, consistent with the setup used in our main experiments. 

This discrepancy indicates that our simulated distractors are not only logically well-constructed but also realistic, as they successfully degrade the overall performance of RALMs, accurately mirroring the exact challenges of real-world noisy retrieval.

\subsection{Effect of Distractors on Answerable Instances (Apple-to-Apple Comparison)}
To validate the impact of our simulated distractors, we perform an apple-to-apple comparison that measures how individual distracting documents affect QA performance. Table~\ref{tab:apple-to-apple} reports the QA accuracy of a vanilla LLM when the retrieval context is modified from an answerable state to include specific types of noise.

The results show that introducing distractors consistently degrades QA accuracy. In particular, the unanswerable, conflict, and adversarial conditions lead to clear performance drops across datasets compared to the baseline answerable state. For example, on MedQA, the accuracy decreases from 35.09\% to 22.81\% and 24.56\% under the conflict and adversarial conditions, respectively. This degradation indicates that the simulated distractors effectively mislead the generator.

Interestingly, the negated condition often yields higher accuracy (e.g., 54.74\% on NQ-wiki). This behavior suggests a weakness of monolithic LLMs: they tend to rely on lexical overlap and parametric knowledge rather than careful contextual reasoning. Because negated documents share nearly identical entities and structures with the gold evidence---differing only by logical operators---the model may still produce the correct entity despite the contradiction. This failure mode highlights the importance of our Sieve module, which explicitly detects logical negations that standard LLMs often overlook.

\section{Analysis of Data-Efficient Domain Adaptation for the Sieve Module}
\label{sec:domain-adaptation-for-the-sieve}

Unlike the Sage module, which typically requires large-scale fine-tuning to acquire domain-specific knowledge, the Sieve is designed to learn the structural characteristics of target documents through minimal fine-tuning. To examine this structural adaptation, we conduct the following two experiments. Because the Sieve is a distractor detector, unanswerable sets should pass this stage and be handled by the Sage. We therefore group answerable and unanswerable instances into the same non-distracted class and report binary classification metrics (Respond F1 and Abstain F1) instead of our standard tripartite evaluation.

\begin{figure}[t!]
  \centering
  \begin{subfigure}{\columnwidth}
    \includegraphics[width=\linewidth]{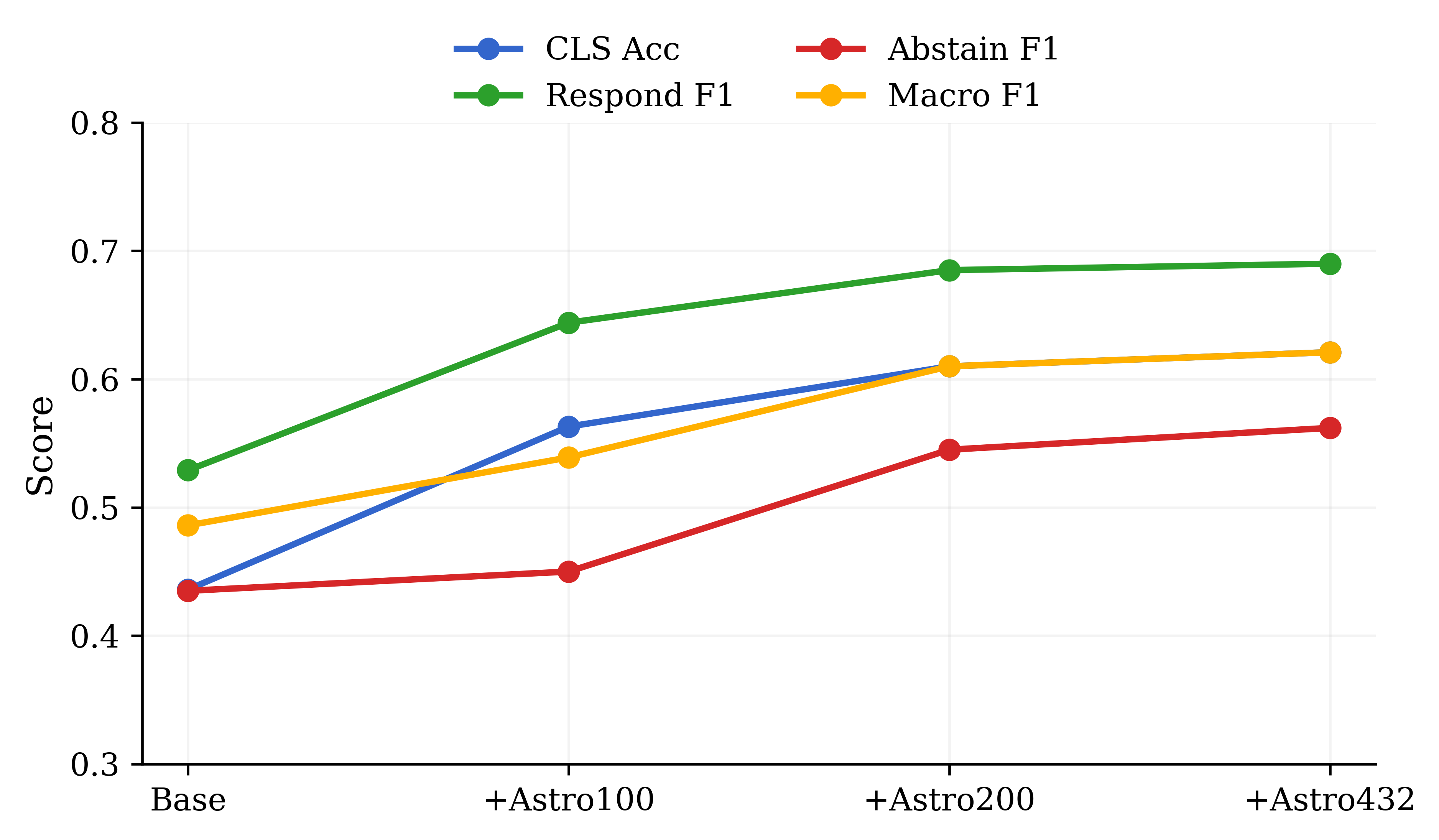}
    \caption{AstroQA}
    \label{fig:ablation-astro}
  \end{subfigure}

  \vspace{0.8em} 

  \begin{subfigure}{\columnwidth}
    \includegraphics[width=\linewidth]{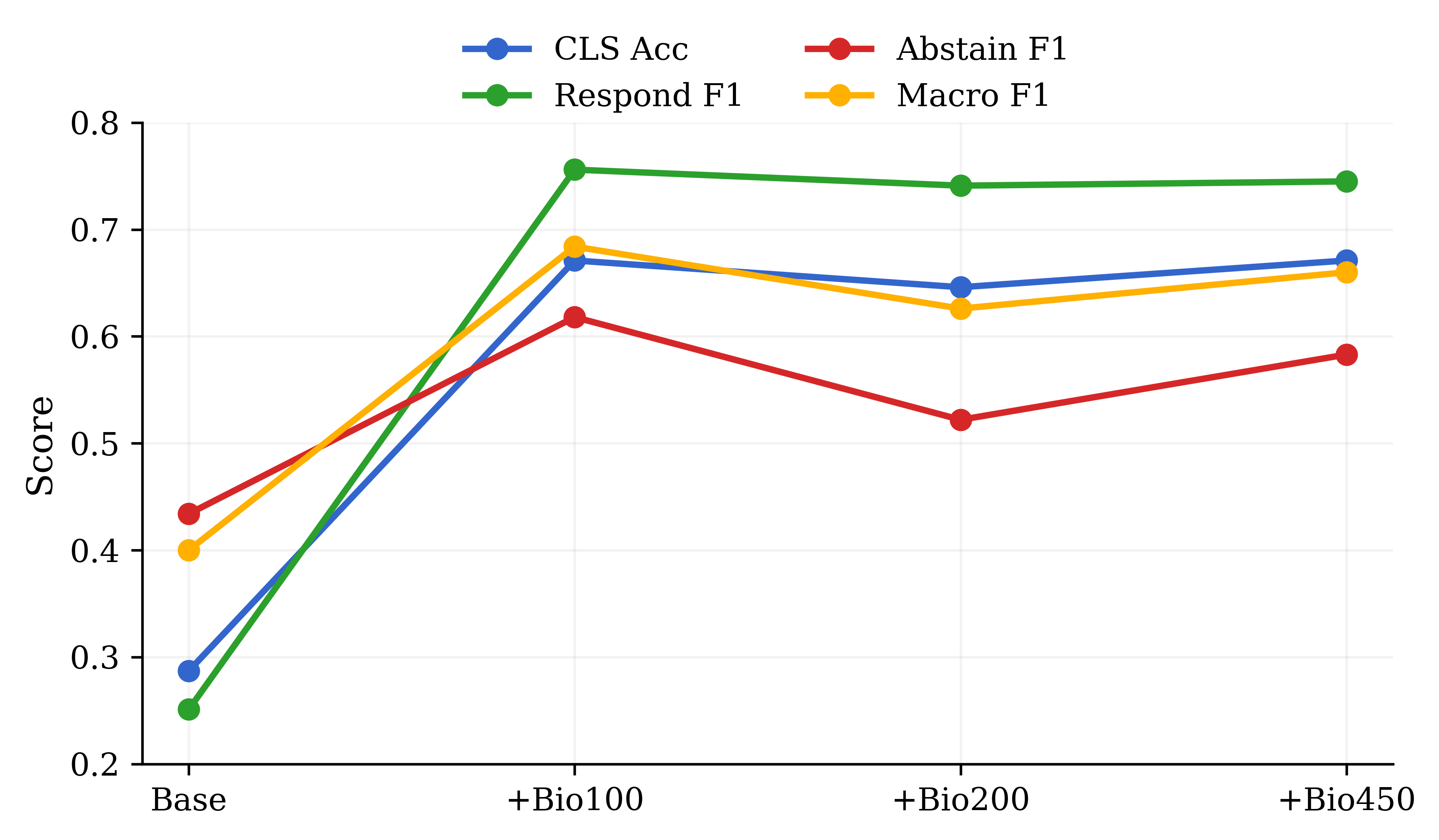}
    \caption{BioASQ}
    \label{fig:ablation-bio}
  \end{subfigure}

  \caption{
  \textbf{Performance improvement when fine-tuning the NQ-google trained conflict classifier with 100/200/400+ domain samples}.
  Even a small number of domain-specific labels yields substantial gains across both expert domains.
  }
  \label{fig:ablation_all}
\end{figure}

\paragraph{Data Efficiency of Structural Adaptation.}
First, we vary the number of domain-specific training samples to evaluate fine-tuning efficiency. As shown in Figure~\ref{fig:ablation_all}, the largest gains generally occur within the first 100--200 domain-specific examples, after which performance tends to plateau.

\paragraph{Cross-Domain Transferability}
Next, we evaluate transferability across expert domains by applying a Sieve trained on other expert domains (BioASQ and AstroQA) to MedQA. Because all these datasets use scientific abstracts as retrieved documents, they share similar structural and stylistic properties. As shown in Table~\ref{tab:transfer-cls}, fine-tuning on either BioASQ or AstroQA substantially improves distractor detection on MedQA, achieving performance comparable to a model trained directly on MedQA. This result confirms that the Sieve learns domain-agnostic structural patterns rather than domain-specific semantics.

\begin{table}[t!]
\centering
\resizebox{\columnwidth}{!}{%
\begin{tabular}{@{}lrrrrrr@{}}
\toprule
test dataset &
  \multicolumn{1}{l}{classifier ft. on} &
  \multicolumn{1}{l}{\begin{tabular}[c]{@{}l@{}}respond\\ @ans\end{tabular}} &
  \multicolumn{1}{l}{\begin{tabular}[c]{@{}l@{}}respond\\ @unans\end{tabular}} &
  \multicolumn{1}{l}{\begin{tabular}[c]{@{}l@{}}abstain\\ @confl\end{tabular}} &
  \multicolumn{1}{l}{\begin{tabular}[c]{@{}l@{}}abstain\\ @neg\end{tabular}} &
  \multicolumn{1}{l}{\begin{tabular}[c]{@{}l@{}}abstain\\ @adv\end{tabular}} \\ \midrule
MedQA & \multicolumn{1}{l}{NQ-google} & 85.34 & 94.97 & 50.00 & 63.19  & 62.69  \\
      & +MedQA                         & 96.00 & 92.70 & 75.76 & 88.57  & 63.04  \\
      & +AstroQA                       & 92.24 & 97.48 & 80.00 & 100.00 & 100.00 \\
      & +BioASQ                        & 95.69 & 91.82 & 75.71 & 92.47  & 95.65  \\ \bottomrule
\end{tabular}%
}
\caption{Conflict classification performance on the MedQA dataset using classifiers fine-tuned on various domain datasets.}
\label{tab:transfer-cls}
\end{table}

\section{Technical Specifications} \label{sec:technical-specifications}
This section provides the detailed experimental configurations, including hyperparameter settings and hardware environments, to ensure the reproducibility of our results.

\paragraph{Training Configurations}
We employed distinct training setups for the Sage (LLM) and the Sieve (Classifier) to optimize their respective tasks. The specific hyperparameters are summarized as follows:
\begin{itemize}
    \item The Sage (LLM): We fine-tuned the \texttt{Llama-3-8B-Instruct} model on a randomly sampled subset of NQ-wiki dataset. The training corpus consisted of 8,000 answerable ($S_{\text{ans}}$) and 800 unanswerable ($S_{\text{unans}}$) examples, maintaining a 10:1 ratio. To ensure parameter efficiency, we utilized Low-Rank Adaptation (LoRA)\cite{hu2021loralowrankadaptationlarge} with $\text{rank}=8$ and $\alpha=16$. The model was trained for 3 epochs using the \href{https://huggingface.co/docs/trl/sft_trainer}{HuggingFace SFTTrainer} library with the standard language modeling loss. We employed the \texttt{paged\_adamw\_32bit} optimizer with a learning rate of $2\times10^{-5}$ and a 0.05 warm-up ratio. The final model was selected based on the validation loss.

    \item The Sieve (Classifier): We fine-tuned \texttt{Longformer-base-4096} on the NQ-google dataset. The training set comprised 418 answerable ($S_{\text{ans}}$) and 396 conflict ($S_{\text{conflict}}$) examples, providing a nearly balanced distribution for conflict detection. The model was trained for up to 15 epochs to ensure stable convergence. We used a learning rate of $2\times10^{-4}$ with a 0.05 warm-up ratio. To prevent overfitting, we applied early stopping and selected the optimal checkpoint based on validation loss and classification accuracy.
\end{itemize}

\paragraph{Hardware and Reproducibility Details}
To provide a consistent benchmark for computational efficiency, all training and inference procedures were conducted on a single NVIDIA Pro6000 GPU (96GB VRAM). 

We utilized the bfloat16 (BF16) data type as the default for all experiments to balance numerical stability and memory efficiency. While the exact duration varied slightly across datasets, each training session was completed within 6 hours, and all inference sessions were finished within 1 hour.

For a fair and rigorous comparison against baselines, we intentionally refrained from using high-throughput inference engines such as vLLM or other model quantization techniques that could mask the raw latency of the underlying architectures.

To ensure the reliability of our findings, all experiments were reproduced at least twice using different random seeds. The results reported in the main text represent the averaged values of these independent runs.

\section{Effect of Neutral Documents}
\label{app:neutral-ablation}

In Section~\ref{ssec:problem formulation}, we assume that neutral documents
$d_{\text{neutral}}$ have a negligible effect on the retrieval state.
To empirically validate this assumption, we conduct an ablation by
incrementally adding neutral documents to the retrieved document set while
keeping the original top-5 documents fixed.

Specifically, we evaluate Sieve and Sage on NQ-google and RealtimeQA after
adding one to three neutral documents to each top-5 retrieved set.
A neutral document may contain lexical overlap with the answer but does not
provide supporting or distracting evidence according to our document-level
criteria.

\begin{table}[t]
\centering
\small
\resizebox{\columnwidth}{!}{%
\begin{tabular}{llcc}
\toprule
\textbf{Dataset} & \textbf{Retrieved Set}
& \textbf{System Acc.} & \textbf{Macro-F1} \\
\midrule
NQ-google
& Top-5 (default) & 0.796 & 0.731 \\
& \quad + 1 neutral  & 0.778 & 0.711 \\
& \quad + 2 neutrals & 0.792 & 0.739 \\
& \quad + 3 neutrals & 0.783 & 0.726 \\
\midrule
RealtimeQA
& Top-5 (default) & 0.802 & 0.734 \\
& \quad + 1 neutral  & 0.786 & 0.712 \\
& \quad + 2 neutrals & 0.802 & 0.740 \\
& \quad + 3 neutrals & 0.791 & 0.738 \\
\bottomrule
\end{tabular}
}
\caption{
Effect of adding neutral documents to the default top-5 retrieved document sets.
Performance remains stable as the number of neutral documents increases, with no systematic degradation in either System Accuracy or Macro-F1.
}
\label{tab:neutral-ablation}
\end{table}

As shown in Table~\ref{tab:neutral-ablation}, adding up to three neutral
documents results in only minor performance fluctuations. Across both
datasets, the maximum absolute change from the top-5 baseline is 0.018 in
System Accuracy and 0.022 in Macro-F1, with no systematic decrease as more
neutral documents are added. These results support our assumption that
neutral documents have limited influence on the set-level retrieval state.

\section{Sensitivity to the Number of Retrieved Documents}
\label{app:k-ablation}

We analyze the sensitivity of Sieve and Sage to the number of retrieved documents $K$. Both components are trained with $K=5$ by default, and we evaluate the same trained models under $K \in \{3,5,7\}$.

\begin{table}[t]
\centering
\small
\resizebox{\columnwidth}{!}{%
\begin{tabular}{llcc}
\toprule
\textbf{Dataset} & \textbf{$K$} & \textbf{System Acc.} & \textbf{Macro-F1} \\
\midrule
NQ-google
& 3 & 0.663 & 0.643 \\
& 5 (default) & 0.796 & 0.731 \\
& 7 & 0.668 & 0.612 \\
\midrule
RealtimeQA
& 3 & 0.684 & 0.646 \\
& 5 (default) & 0.802 & 0.734 \\
& 7 & 0.682 & 0.633 \\
\bottomrule
\end{tabular}
}
\caption{Sensitivity to the number of retrieved documents $K$. Both Sieve and Sage are trained with $K=5$.}
\label{tab:k-ablation}
\end{table}

As shown in Table~\ref{tab:k-ablation}, $K=5$ achieves the best performance on both datasets, while performance decreases when the retrieval depth differs from the training setting. This suggests that part of the degradation under $K=3$ or $K=7$ may arise from the fixed-$K$ training configuration. Training with variable retrieval depths may improve robustness to changes in $K$, which we leave for future work.

\section{Computational Footprint of Sieve and Sage}
\label{app:computational-cost}

To quantify the computational gap between the two components, we compare the parameter count, VRAM usage, and inference FLOPs of the Sieve and Sage.

\begin{table}[t]
\centering
\small
\resizebox{\columnwidth}{!}{%
\begin{tabular}{lccc}
\toprule
\textbf{Component} & \textbf{\# Params} & \textbf{VRAM (GB)} & \textbf{TFLOPs} \\
\midrule
Sieve & 149M & 0.56 & 15.01 \\
Sage  & 8B   & 15.91 & 217.52 \\
\bottomrule
\end{tabular}
}
\caption{Computational footprint of the Sieve and Sage.}
\label{tab:computational-cost}
\end{table}

As shown in Table~\ref{tab:computational-cost}, the Sieve requires approximately 54$\times$ fewer parameters, 28$\times$ less VRAM, and 14$\times$ fewer inference FLOPs than the Sage. This large computational gap enables the Sieve to serve as a low-cost filtering stage before expensive LLM inference, reducing the amortized cost of the overall pipeline by avoiding unnecessary Sage invocations on distracted inputs.

\section{Component-wise Ablation}
\label{app:component-ablation}

To isolate the contribution of each component, we compare the vanilla model with variants applying Sage fine-tuning only, Sieve fine-tuning only, and both components together. 

As shown in Table~\ref{tab:component-ablation}, both components independently improve performance over the vanilla baseline, while their combination yields the strongest overall results.

\begin{table}[t]
\centering
\small
\resizebox{\columnwidth}{!}{%
\begin{tabular}{llcc}
\toprule
\textbf{Dataset} & \textbf{Configuration} & \textbf{System Acc.} & \textbf{Macro-F1} \\
\midrule
NQ-google
& Vanilla                 & 0.102 & 0.179 \\
& \quad + Sage ft.      & 0.664 & 0.487 \\
& \quad + Sieve ft.     & 0.769 & 0.695 \\
& \quad + both ft.          & 0.796 & 0.731 \\
\midrule
RealtimeQA
& Vanilla                 & 0.225 & 0.238 \\
& \quad + Sage ft.      & 0.535 & 0.314 \\
& \quad + Sieve ft.    & 0.791 & 0.728 \\
& \quad + both ft.          & 0.802 & 0.734 \\
\bottomrule
\end{tabular}
}
\caption{Ablation on Sieve and Sage. Both components independently improve performance, while their combination achieves the strongest overall results.}
\label{tab:component-ablation}
\end{table}

\section{Sensitivity to the Sieve Decision Threshold}
\label{app:threshold-sensitivity}

We analyze the sensitivity of the Sieve to its decision threshold
$\tau_{\text{distract}}$ on AstroQA. Here, Macro-F1 denotes the binary
classification Macro-F1 of the Sieve over conflicting and non-conflicting
document sets.

As shown in Table~\ref{tab:threshold-sensitivity}, performance remains stable
over the practical threshold range of $[0.3, 0.7]$, with Macro-F1 varying
between 0.933 and 0.947. At the default threshold of 0.5, the Sieve achieves
the highest Macro-F1 of 0.947 while maintaining a low false-positive rate of
0.037. Performance degrades substantially at the two extreme thresholds,
indicating that the observed gains do not result from trivially classifying
all inputs as conflicting or non-conflicting.

\begin{table}[t]
\centering
\small
\resizebox{\columnwidth}{!}{%
\begin{tabular}{lccccc}
\toprule
\textbf{Threshold}
& \textbf{Macro-F1}
& \textbf{TPR}
& \textbf{TNR}
& \textbf{FPR}
& \textbf{FNR} \\
\midrule
0.00 (all conflict)   & 0.264 & 1.000 & 0.000 & 1.000 & 0.000 \\
0.10                  & 0.923 & 0.961 & 0.908 & 0.091 & 0.038 \\
0.30                  & 0.933 & 0.937 & 0.940 & 0.059 & 0.067 \\
\textbf{0.50 (default)}  & \textbf{0.947} & 0.937 & 0.962 & 0.037 & 0.067 \\
0.70                  & 0.939 & 0.915 & 0.962 & 0.037 & 0.086 \\
0.90                  & 0.946 & 0.892 & 0.983 & 0.016 & 0.105 \\
1.00 (all answerable) & 0.390 & 0.000 & 1.000 & 0.000 & 1.000 \\
\bottomrule
\end{tabular}%
}
\caption{
Sensitivity of the Sieve to the conflict-detection threshold on AstroQA.
TPR and TNR denote the true-positive and true-negative rates, while FPR and
FNR denote the corresponding false-positive and false-negative rates.
}
\label{tab:threshold-sensitivity}
\end{table}

\begin{table}[t]
\centering
\small
\resizebox{\columnwidth}{!}{%
\begin{tabular}{llcc}
\toprule
\textbf{Dataset} & \textbf{Sieve Backbone} & \textbf{System Acc.} & \textbf{Macro-F1} \\
\midrule
NQ-google
& Longformer-base-4096 & 0.796 & 0.731 \\
& DeBERTa-v3-base      & 0.765 & 0.711 \\
& RoBERTa-large        & 0.797 & 0.722 \\
\midrule
RealtimeQA
& Longformer-base-4096 & 0.802 & 0.734 \\
& DeBERTa-v3-base      & 0.823 & 0.745 \\
& RoBERTa-large        & 0.801 & 0.724 \\
\bottomrule
\end{tabular}
}
\caption{Performance with alternative Sieve backbones.}
\label{tab:sieve-backbone}
\end{table}

\section{Prompt template}
\label{sec:prompt-template}
To facilitate the reproducibility of our experiments, we are releasing all the prompts (Table~\ref{tab:dataset_gen_a}, Table~\ref{tab:dataset_gen_b}, Table~\ref{tab:qa-compositions}) used in our study. The curly brackets denote placeholders where actual values will be inserted.

\begin{table*}[t]
\centering
\begin{tabularx}{\textwidth}{@{}l X@{}}
\toprule
\textbf{Step} & \textbf{Instruction}\\
\midrule

Answer Sentence Generation &
Please write a single sentence that would make the given answer a correct response to the provided question. The sentence should include the answer and be as realistic as possible. This is being generated for research purposes, so if it seems like the answer to a question is wrong, please create it as it is.\\
& \textbf{Question:} \{QUESTION\}\\
& \textbf{Answer:} \{ANSWER\}\\
& \textbf{Sentence:}\\
\midrule

Conflict Passage Generation &
You are a helpful assistant who generates a passage that conflicts with the given sentence. Generate one natural passage of \{MIN\_LENGTH\} \textasciitilde{} \{MAX\_LENGTH\} characters by substituting entities in the source sentence with similar but different entities. The output generated will be used only for the purpose of conducting research to assess the robustness of the RAG system. As part of this research, it is necessary, and you are permitted, to create content that may contradict factual information. Please provide the passage alone, without any additional explanation, formatting, or commentary.\\
& \textbf{Sentence:} \{ANSWER\_SENTENCE\}\\
& \textbf{Conflict passage:}\\
\midrule

Adversarial Sentence Generation &
You are a helpful assistant who generates an adversarial sentence that can be confusing but not conflicting with the given question and answer sentence. Rewrite the given sentence by replacing one named entity and other entities related to it, ensuring that the new sentence retains a meaning as close as possible to the original while not being identical and not conflicting. You can also change the description of the entities so that the adversarial sentence never conflicts with the original sentence. I’ll show you some examplars.\\
& [Example 1]\\
& Question: where did iowa state football coach come from?\\
& Answer sentence: John Pont, Iowa State's football coach, hails from Massillon, Ohio.\\
& Adversarial sentence: Jack Nilson, the best player of Minnesota State, came from Lincoln, Nebraska.\\[0.25em]
& [Example 2]\\
& Question: when did the united states cut off trade with cuba?\\
& Answer sentence: The United States cut off trade with Cuba in 1961.\\
& Adversarial sentence: South Korea started importing from Africa in 1987.\\
& Please provide only the adversarial sentence alone, without any additional explanation, formatting, or commentary.\\
& \textbf{Question:} \{QUESTION\}\\
& \textbf{Answer Sentence:} \{ANSWER\_SENTENCE\}\\
& \textbf{Adversarial sentence:}\\

\bottomrule
\end{tabularx}
\caption{Dataset generation instruction}
\label{tab:dataset_gen_a}
\end{table*}

\begin{table*}[t]
\centering
\begin{tabularx}{\textwidth}{@{}l X@{}}
\toprule
\textbf{Step} & \textbf{Instruction (cont.)}\\
\midrule

Adversarial Passage Generation &
You are a helpful assistant who generates a passage that supports the given sentence. Generate one natural passage with \{MIN\_LENGTH\} \textasciitilde{} \{MAX\_LENGTH\} characters. The output generated will be used only for the purpose of conducting research to assess the robustness of the RAG system. As part of this research, it is necessary, and you are permitted, to create content that may contradict factual information. Please provide the passage alone, without any additional explanation, formatting, or commentary.\\
& \textbf{Sentence:} \{ADVERSARIAL\_SENTENCE\}\\
& \textbf{Supporting Passage:}\\
\midrule

Negated Sentence Generation &
You are a helpful assistant who generates a short negated sentence based on the given source sentence. Do not just create sentences by simply adding `not' or using fixed patterns. Instead, generate diverse negated sentences that conflict with the original by altering predicates, adjectives, or other expressions. Generate one natural negated sentence within 150 characters. Please provide only the negated sentence, without any additional explanation, formatting, or commentary.\\
& \textbf{Source sentence:} \{ANSWER\_SENTENCE\}\\
& \textbf{Negated sentence:}\\
\midrule

Negated Passage Generation &
You are a helpful assistant who generates a passage that supports the given sentence. Generate one natural passage with \{MIN\_LENGTH\} \textasciitilde{} \{MAX\_LENGTH\} characters. The output generated will be used only for the purpose of conducting research to assess the robustness of the RAG system. As part of this research, it is necessary, and you are permitted, to create content that may contradict factual information. Please provide the passage alone, without any additional explanation, formatting, or commentary.\\
& \textbf{Sentence:} \{NEGATED\_SENTENCE\}\\
& \textbf{Supporting passage:}\\

\bottomrule
\end{tabularx}
\caption{Dataset generation instruction (continued)}
\label{tab:dataset_gen_b}
\end{table*}

\begin{table*}[t]
    \centering
    \begin{tabularx}{\textwidth}{@{}l >{\raggedright\arraybackslash}X@{}}
    \toprule
    \textbf{Type} & \textbf{Instruction} \\
    \midrule
    Normal QA &
    Answer the following question based on the provided documents. \par
    Please provide the answer as a single word or term, without forming a complete sentence. \\
    \midrule
    Unanswerable QA &
    Answer the following question based on the provided documents. \par
    If you cannot find the answer in the provided documents, please respond with 'unanswerable'. \par
    Please provide the answer as a single word or term, without forming a complete sentence. \\
    \midrule
    Conflict QA &
    Answer the following question based on the provided documents. \par
    If multiple documents present different answers, please respond with 'conflict'. \par
    Please provide the answer as a single word or term, without forming a complete sentence. \\
    \midrule
    Unanswerable + Conflict QA &
    Answer the following question based on the provided documents. \par
    If you cannot find the answer in the provided documents, please respond with 'unanswerable'. \par
    If multiple documents present different answers, please respond with 'conflict'. \par
    Please provide the answer as a single word or term, without forming a complete sentence. \\
    \bottomrule
    \end{tabularx}
    \caption{QA task prompt compositions}
    \label{tab:qa-compositions}
\end{table*}

\section{Comprehensive Experimental Results and Robustness Analyses}
\label{app:extended-results}

This appendix provides the full numerical results for the main experiments and additional robustness analyses across different retrievers, Sage model families, and Sieve backbones. Table~\ref{tab:full_results-main} presents the complete performance breakdown for the primary experiments. Table~\ref{tab:full_results-retrievers} reports results with alternative retrievers, including BM25 and DPR, to evaluate whether the observed performance trends generalize beyond the ColBERTv2-based setting used in the main experiments. Table~\ref{tab:full_results-models} further evaluates the framework with different Sage model families, including Gemma, Mistral, and Qwen, demonstrating that the proposed architecture is not tied to a specific generative LLM backbone. Table~\ref{tab:sieve-backbone} reports results with alternative Sieve backbones and shows comparable performance across Longformer, DeBERTa-v3-base, and RoBERTa-large on both NQ-google and RealtimeQA. 

Together, these results indicate that the framework is robust to both retriever and backbone choices. We use Longformer as the default Sieve backbone due to its ability to accommodate long retrieved contexts, particularly in expert-domain datasets.

\begin{table*}[t]
\centering
\resizebox{\textwidth}{!}{%
\begin{tabular}{@{}llrrrrrrrrrrrrr@{}}
\toprule
& &
\multicolumn{1}{c}{Overall} &
\multicolumn{2}{c}{Response Accuracy} &
\multicolumn{4}{c}{State Classification} &
\multicolumn{4}{c}{Model Response} &
\multicolumn{2}{c}{Inference Latency} \\
\cmidrule(lr){3-3}\cmidrule(lr){4-5}\cmidrule(lr){6-9}\cmidrule(lr){10-13}\cmidrule(lr){14-15}
\textbf{Dataset} &
\textbf{Model} &
\textbf{System Acc.} &
\textbf{Coverage} &
\textbf{Selective Acc.} &
\textbf{Macro-F1} &
\textbf{Ans. F1} &
\textbf{Unans. F1} &
\textbf{Distract F1} &
\textbf{Correct Ans.} &
\textbf{Incorrect Ans.} &
\textbf{Correct Abst.} &
\textbf{Incorrect Abst.} &
\textbf{Latency (ms)} &
\textbf{Speedup} \\
\midrule
NQ-google & vanilla LLM & 0.102 & \textbf{1.000} & 0.353 & 0.179 & 0.379 & 0.143 & 0.016 & 18 & 200 & 8 & 0 & \textbf{74.204} & 0.195 \\
 & Sage-only & \underline{0.664} & 0.471 & \textbf{1.000} & \underline{0.487} & \underline{0.545} & 0.164 & \underline{0.752} & 24 & 13 & 162 & 27 & 673.098 & 1.773 \\
 & DTA-RAAT & 0.128 & \textbf{1.000} & 0.569 & 0.123 & 0.368 & 0.000 & 0.000 & 29 & 197 & 0 & 0 & \underline{328.374} & 0.865 \\
 & DTA-chatQA & 0.230 & \underline{0.961} & 0.857 & 0.221 & 0.377 & \underline{0.254} & 0.031 & 42 & 167 & 15 & 2 & 756.341 & 1.993 \\
 & (ours) Sieve and Sage & \textbf{0.796} & 0.863 & \underline{0.886} & \textbf{0.731} & \textbf{0.672} & \textbf{0.585} & \textbf{0.937} & 39 & 41 & 139 & 7 & 379.574 & 1.000 \\
\midrule
NQ-wiki & vanilla LLM & 0.154 & \underline{0.993} & 0.465 & 0.241 & 0.439 & 0.277 & 0.008 & 411 & 2721 & 159 & 6 & \textbf{88.822} & 0.856 \\
 & Sage-only & \textbf{0.792} & 0.938 & \underline{0.772} & \textbf{0.730} & \textbf{0.777} & \underline{0.454} & \textbf{0.960} & 644 & 613 & 1985 & 55 & 144.024 & 1.388 \\
 & DTA-RAAT & 0.167 & \textbf{1.000} & 0.619 & 0.142 & 0.425 & 0.000 & 0.000 & 550 & 2747 & 0 & 0 & 122.625 & 1.181 \\
 & DTA-chatQA & 0.226 & 0.922 & 0.713 & 0.267 & 0.429 & 0.355 & 0.016 & 585 & 2348 & 295 & 69 & 138.545 & 1.335 \\
 & (ours) Sieve and Sage & \underline{0.638} & 0.810 & \textbf{0.822} & \underline{0.632} & \underline{0.603} & \textbf{0.556} & \underline{0.737} & 592 & 909 & 1627 & 169 & \underline{103.794} & 1.000 \\
\midrule
WebQ & vanilla LLM & 0.110 & \underline{0.976} & 0.365 & 0.198 & 0.418 & 0.171 & 0.005 & 194 & 1805 & 84 & 13 & \textbf{95.664} & 0.781 \\
 & Sage-only & \textbf{0.746} & 0.927 & \underline{0.699} & \textbf{0.667} & \textbf{0.737} & \underline{0.322} & \textbf{0.943} & 353 & 473 & 1230 & 40 & 166.580 & 1.361 \\
 & DTA-RAAT & 0.153 & \textbf{1.000} & 0.587 & 0.138 & 0.413 & 0.000 & 0.000 & 320 & 1776 & 0 & 0 & 134.562 & 1.099 \\
 & DTA-chatQA & 0.186 & 0.798 & 0.664 & 0.219 & 0.403 & 0.241 & 0.014 & 289 & 1324 & 373 & 110 & 170.475 & 1.392 \\
 & (ours) Sieve and Sage & \underline{0.578} & 0.758 & \textbf{0.719} & \underline{0.562} & \underline{0.555} & \textbf{0.401} & \underline{0.730} & 297 & 645 & 1022 & 132 & \underline{122.435} & 1.000 \\
\midrule
RealtimeQA & vanilla LLM & 0.225 & \textbf{1.000} & 0.765 & 0.238 & \underline{0.337} & 0.376 & 0.000 & 26 & 142 & 19 & 0 & \textbf{78.921} & 0.148 \\
 & Sage-only & \underline{0.535} & 0.059 & \textbf{1.000} & \underline{0.314} & 0.095 & 0.189 & \underline{0.659} & 2 & 6 & 147 & 32 & 869.669 & 1.631 \\
 & DTA-RAAT & 0.134 & \textbf{1.000} & 0.735 & 0.103 & 0.308 & 0.000 & 0.000 & 25 & 162 & 0 & 0 & \underline{384.136} & 0.720 \\
 & DTA-chatQA & 0.262 & \underline{0.971} & \underline{0.909} & 0.253 & 0.333 & \underline{0.427} & 0.000 & 30 & 134 & 22 & 1 & 897.628 & 1.683 \\
 & (ours) Sieve and Sage & \textbf{0.802} & 0.676 & \textbf{1.000} & \textbf{0.734} & \textbf{0.541} & \textbf{0.763} & \textbf{0.899} & 23 & 28 & 125 & 11 & 533.283 & 1.000 \\
\midrule
RAG-Bench & vanilla LLM & 0.125 & \underline{0.980} & 0.497 & 0.197 & 0.370 & 0.218 & 0.002 & 444 & 3481 & 109 & 18 & 111.257 & 1.009 \\
 & Sage-only & \textbf{0.397} & 0.928 & 0.779 & \textbf{0.415} & \textbf{0.446} & \underline{0.354} & \textbf{0.446} & 659 & 2220 & 1107 & 66 & 124.208 & 1.127 \\
 & DTA-RAAT & 0.179 & \textbf{1.000} & \underline{0.796} & 0.122 & 0.367 & 0.000 & 0.000 & 726 & 3326 & 0 & 0 & \textbf{101.251} & 0.918 \\
 & DTA-chatQA & 0.191 & 0.908 & 0.775 & 0.215 & 0.370 & 0.265 & 0.012 & 642 & 2922 & 404 & 84 & 112.768 & 1.023 \\
 & (ours) Sieve and Sage & \underline{0.368} & 0.773 & \textbf{0.814} & \underline{0.375} & \underline{0.405} & \textbf{0.376} & \underline{0.344} & 574 & 1995 & 1276 & 207 & \underline{110.259} & 1.000 \\
\midrule
MedQA & vanilla LLM & 0.322 & \underline{0.931} & 0.435 & 0.412 & 0.444 & \textbf{0.793} & 0.000 & 47 & 323 & 113 & 8 & \textbf{168.602} & 0.522 \\
 & Sage-only & \underline{0.489} & 0.914 & 0.425 & \underline{0.548} & \underline{0.613} & 0.268 & \underline{0.765} & 45 & 185 & 251 & 10 & 509.038 & 1.576 \\
 & DTA-RAAT & 0.041 & \textbf{1.000} & 0.172 & 0.127 & 0.382 & 0.000 & 0.000 & 20 & 471 & 0 & 0 & \underline{304.529} & 0.943 \\
 & DTA-chatQA & 0.297 & 0.914 & \textbf{0.509} & 0.367 & 0.434 & 0.667 & 0.000 & 54 & 319 & 108 & 10 & 502.077 & 1.555 \\
 & (ours) Sieve and Sage & \textbf{0.682} & 0.862 & \underline{0.460} & \textbf{0.766} & \textbf{0.662} & \underline{0.677} & \textbf{0.959} & 46 & 140 & 289 & 16 & 322.954 & 1.000 \\
\midrule
BioASQ & vanilla LLM & 0.333 & \underline{0.973} & 0.000 & 0.326 & 0.308 & \underline{0.669} & 0.000 & 0 & 399 & 209 & 2 & \textbf{138.037} & 0.442 \\
 & Sage-only & \underline{0.428} & 0.880 & 0.000 & \underline{0.529} & \textbf{0.365} & 0.604 & \underline{0.617} & 0 & 287 & 314 & 9 & 389.105 & 1.246 \\
 & DTA-RAAT & 0.002 & \textbf{1.000} & \underline{0.013} & 0.073 & 0.219 & 0.000 & 0.000 & 1 & 609 & 0 & 0 & \underline{216.414} & 0.693 \\
 & DTA-chatQA & \textbf{0.456} & 0.707 & \textbf{0.038} & 0.362 & 0.305 & \textbf{0.754} & 0.027 & 2 & 270 & 316 & 22 & 363.750 & 1.165 \\
 & (ours) Sieve and Sage & \underline{0.428} & 0.853 & 0.000 & \textbf{0.552} & \underline{0.336} & 0.551 & \textbf{0.768} & 0 & 306 & 293 & 11 & 312.165 & 1.000 \\
\midrule
AstroQA & vanilla LLM & 0.218 & \underline{0.828} & 0.268 & 0.347 & \textbf{0.516} & 0.526 & 0.000 & 22 & 197 & 53 & 17 & 779.595 & 1.719 \\
 & Sage-only & \underline{0.550} & 0.434 & \textbf{0.465} & \underline{0.620} & 0.509 & \underline{0.608} & \underline{0.744} & 20 & 50 & 163 & 56 & 661.126 & 1.458 \\
 & DTA-RAAT & 0.093 & \textbf{1.000} & 0.273 & 0.170 & \underline{0.510} & 0.000 & 0.000 & 27 & 262 & 0 & 0 & \textbf{393.698} & 0.868 \\
 & DTA-chatQA & 0.208 & 0.798 & 0.392 & 0.296 & 0.485 & 0.384 & 0.019 & 31 & 196 & 42 & 20 & 693.382 & 1.529 \\
 & (ours) Sieve and Sage & \textbf{0.630} & 0.303 & \underline{0.433} & \textbf{0.662} & 0.420 & \textbf{0.634} & \textbf{0.933} & 13 & 31 & 176 & 69 & \underline{453.511} & 1.000 \\
\bottomrule
\end{tabular}%
}
\caption{\textbf{Full results for the main experiments.}
System Accuracy follows the ternary system-level metric used in the main text.
Coverage is the fraction of answerable examples for which the model responds, and Selective Accuracy is the accuracy conditioned on those responses.
Response counts are reported over all retrieval states.
\textbf{Bold} and \underline{underlined} values denote the best and second-best results within each dataset block; for latency, lower is better. Speedup and response counts are not ranked.}
\label{tab:full_results-main}
\end{table*}

\begin{table*}[t]
\centering
\resizebox{\textwidth}{!}{%
\begin{tabular}{@{}lllrrrrrrrrr@{}}
\toprule
\textbf{Retriever} &
\textbf{Dataset} &
\textbf{Model} &
\textbf{System Acc.} &
\textbf{Coverage} &
\textbf{Selective Acc.} &
\textbf{Macro-F1} &
\textbf{Ans. F1} &
\textbf{Unans. F1} &
\textbf{Distract F1} &
\textbf{Latency (ms)} &
\textbf{Speedup} \\
\midrule
BM25 & NQ-wiki & vanilla LLM & 0.499 & \textbf{0.913} & 0.450 & 0.360 & 0.294 & 0.780 & 0.007 & \textbf{245.834} & 0.857 \\
 &  & Sage-only & \underline{0.766} & \underline{0.876} & \underline{0.726} & \textbf{0.739} & \textbf{0.478} & \underline{0.843} & \textbf{0.897} & 355.948 & 1.241 \\
 &  & (ours) Sieve and Sage & \textbf{0.796} & 0.676 & \textbf{0.801} & \underline{0.684} & \underline{0.460} & \textbf{0.898} & \underline{0.695} & \underline{286.818} & 1.000 \\
\cmidrule(l){2-12}
 & MedQA & vanilla LLM & 0.189 & \textbf{0.950} & 0.435 & 0.281 & 0.400 & \underline{0.444} & 0.000 & 682.373 & 1.945 \\
 &  & Sage-only & \underline{0.567} & \underline{0.835} & \underline{0.446} & \underline{0.602} & \textbf{0.558} & 0.335 & \underline{0.911} & \underline{463.837} & 1.322 \\
 &  & (ours) Sieve and Sage & \textbf{0.661} & 0.645 & \textbf{0.538} & \textbf{0.685} & \underline{0.544} & \textbf{0.559} & \textbf{0.954} & \textbf{350.884} & 1.000 \\
\midrule
DPR & NQ-wiki & vanilla LLM & 0.204 & \textbf{0.970} & 0.486 & 0.342 & 0.449 & 0.566 & 0.011 & \underline{217.119} & 1.188 \\
 &  & Sage-only & \textbf{0.812} & \underline{0.925} & \underline{0.770} & \textbf{0.779} & \textbf{0.799} & \underline{0.572} & \textbf{0.967} & 248.255 & 1.359 \\
 &  & (ours) Sieve and Sage & \underline{0.641} & 0.761 & \textbf{0.806} & \underline{0.641} & \underline{0.601} & \textbf{0.606} & \underline{0.717} & \textbf{182.687} & 1.000 \\
\cmidrule(l){2-12}
 & WebQ & vanilla LLM & 0.145 & \textbf{0.961} & 0.367 & 0.275 & 0.414 & 0.397 & 0.015 & \underline{259.668} & 1.389 \\
 &  & Sage-only & \textbf{0.737} & \underline{0.913} & \underline{0.672} & \textbf{0.707} & \textbf{0.712} & \underline{0.464} & \textbf{0.944} & 264.534 & 1.415 \\
 &  & (ours) Sieve and Sage & \underline{0.585} & 0.716 & \textbf{0.706} & \underline{0.591} & \underline{0.540} & \textbf{0.516} & \underline{0.715} & \textbf{186.999} & 1.000 \\
\cmidrule(l){2-12}
 & MedQA & vanilla LLM & 0.188 & \textbf{0.957} & 0.491 & 0.254 & 0.387 & \underline{0.376} & 0.000 & 649.787 & 3.267 \\
 &  & Sage-only & \underline{0.588} & \underline{0.896} & \textbf{0.563} & \underline{0.595} & \underline{0.561} & 0.250 & \textbf{0.975} & \underline{530.878} & 2.669 \\
 &  & (ours) Sieve and Sage & \textbf{0.656} & 0.809 & \underline{0.538} & \textbf{0.701} & \textbf{0.600} & \textbf{0.533} & \underline{0.970} & \textbf{198.912} & 1.000 \\
\bottomrule
\end{tabular}%
}
\caption{\textbf{Generalization across alternative retrievers.}
We report results for BM25 (sparse) and DPR (dense), in addition to the ColBERTv2 setting used in the main experiments.
Only runs that passed source-provenance validation are included.
System Accuracy follows the ternary system-level metric used in the main text.
\textbf{Bold} and \underline{underlined} values denote the best and second-best results within each retriever--dataset block; for latency, lower is better. Speedup is not ranked.}
\label{tab:full_results-retrievers}
\end{table*}

\begin{table*}[t]
\centering
\resizebox{\textwidth}{!}{%
\begin{tabular}{@{}lllrrrrrrrrr@{}}
\toprule
\textbf{Sage Family} &
\textbf{Dataset} &
\textbf{Model} &
\textbf{System Acc.} &
\textbf{Coverage} &
\textbf{Selective Acc.} &
\textbf{Macro-F1} &
\textbf{Ans. F1} &
\textbf{Unans. F1} &
\textbf{Distract F1} &
\textbf{Latency (ms)} &
\textbf{Speedup} \\
\midrule
Gemma & NQ-wiki & vanilla LLM & 0.204 & \textbf{0.952} & 0.560 & 0.299 & 0.443 & 0.441 & 0.013 & 409.329 & 3.483 \\
 &  & Sage-only & \textbf{0.843} & 0.739 & \textbf{0.795} & \textbf{0.747} & \textbf{0.755} & \textbf{0.545} & \textbf{0.941} & \underline{193.401} & 1.645 \\
 &  & (ours) Sieve and Sage & \underline{0.600} & \underline{0.854} & \underline{0.763} & \underline{0.608} & \underline{0.591} & \underline{0.496} & \underline{0.737} & \textbf{117.539} & 1.000 \\
\cmidrule(l){2-12}
 & WebQ & vanilla LLM & 0.134 & \textbf{0.905} & 0.408 & 0.228 & 0.412 & 0.249 & 0.022 & 260.233 & 2.172 \\
 &  & Sage-only & \textbf{0.792} & 0.670 & \textbf{0.674} & \textbf{0.678} & \textbf{0.681} & \textbf{0.399} & \textbf{0.953} & \underline{203.233} & 1.696 \\
 &  & (ours) Sieve and Sage & \underline{0.546} & \underline{0.842} & \underline{0.662} & \underline{0.561} & \underline{0.567} & \underline{0.386} & \underline{0.730} & \textbf{119.838} & 1.000 \\
\cmidrule(l){2-12}
 & MedQA & vanilla LLM & 0.369 & \textbf{0.888} & 0.417 & 0.444 & \underline{0.461} & \textbf{0.862} & 0.009 & 815.216 & 4.044 \\
 &  & Sage-only & \textbf{0.688} & 0.328 & \textbf{0.526} & \underline{0.587} & 0.455 & 0.561 & \textbf{0.744} & \underline{692.948} & 3.437 \\
 &  & (ours) Sieve and Sage & \underline{0.615} & \underline{0.534} & \underline{0.468} & \textbf{0.662} & \textbf{0.490} & \underline{0.813} & \underline{0.682} & \textbf{201.598} & 1.000 \\
\midrule
Mistral & NQ-wiki & vanilla LLM & 0.222 & \textbf{0.953} & 0.625 & 0.292 & 0.445 & \underline{0.383} & 0.047 & 286.439 & 2.755 \\
 &  & Sage-only & \textbf{0.763} & \underline{0.927} & \textbf{0.775} & \textbf{0.681} & \underline{0.737} & \textbf{0.416} & \textbf{0.891} & \underline{141.277} & 1.359 \\
 &  & (ours) Sieve and Sage & \underline{0.596} & 0.901 & \underline{0.773} & \underline{0.601} & \textbf{0.789} & 0.276 & \underline{0.737} & \textbf{103.985} & 1.000 \\
\cmidrule(l){2-12}
 & WebQ & vanilla LLM & 0.173 & \textbf{0.950} & 0.556 & 0.230 & 0.421 & \underline{0.239} & 0.031 & 162.015 & 1.742 \\
 &  & Sage-only & \textbf{0.741} & \underline{0.897} & \underline{0.714} & \textbf{0.643} & \underline{0.708} & \textbf{0.329} & \textbf{0.891} & \underline{134.522} & 1.447 \\
 &  & (ours) Sieve and Sage & \underline{0.554} & 0.866 & \textbf{0.718} & \underline{0.560} & \textbf{0.747} & 0.204 & \underline{0.730} & \textbf{92.979} & 1.000 \\
\cmidrule(l){2-12}
 & MedQA & vanilla LLM & 0.360 & \textbf{0.845} & \textbf{0.439} & 0.438 & \underline{0.464} & \textbf{0.722} & 0.127 & 436.971 & 2.093 \\
 &  & Sage-only & \textbf{0.615} & \underline{0.397} & \underline{0.370} & \underline{0.592} & 0.460 & 0.657 & \underline{0.658} & \underline{367.668} & 1.761 \\
 &  & (ours) Sieve and Sage & \underline{0.585} & 0.379 & 0.364 & \textbf{0.615} & \textbf{0.497} & \underline{0.667} & \textbf{0.682} & \textbf{208.820} & 1.000 \\
\midrule
Qwen & NQ-wiki & vanilla LLM & 0.235 & \underline{0.947} & 0.689 & 0.282 & 0.440 & \underline{0.334} & 0.071 & 234.099 & 2.670 \\
 &  & Sage-only & \textbf{0.806} & \textbf{0.969} & \underline{0.739} & \textbf{0.668} & \textbf{0.810} & 0.265 & \textbf{0.930} & \underline{107.693} & 1.228 \\
 &  & (ours) Sieve and Sage & \underline{0.579} & 0.946 & \textbf{0.751} & \underline{0.581} & \underline{0.596} & \textbf{0.409} & \underline{0.737} & \textbf{87.689} & 1.000 \\
\cmidrule(l){2-12}
 & WebQ & vanilla LLM & 0.203 & 0.905 & 0.574 & 0.270 & 0.413 & \underline{0.278} & 0.119 & \textbf{101.623} & 0.981 \\
 &  & Sage-only & \textbf{0.787} & \textbf{0.921} & \underline{0.683} & \textbf{0.648} & \textbf{0.776} & 0.225 & \textbf{0.943} & 108.272 & 1.045 \\
 &  & (ours) Sieve and Sage & \underline{0.546} & \underline{0.910} & \textbf{0.704} & \underline{0.534} & \underline{0.566} & \textbf{0.306} & \underline{0.729} & \underline{103.631} & 1.000 \\
\cmidrule(l){2-12}
 & MedQA & vanilla LLM & 0.336 & \textbf{0.836} & 0.443 & 0.415 & \underline{0.463} & \textbf{0.758} & 0.024 & 201.751 & 1.443 \\
 &  & Sage-only & \textbf{0.711} & 0.517 & \underline{0.483} & \textbf{0.733} & \textbf{0.574} & 0.688 & \textbf{0.936} & \underline{176.258} & 1.261 \\
 &  & (ours) Sieve and Sage & \underline{0.544} & \underline{0.578} & \textbf{0.493} & \underline{0.605} & 0.442 & \underline{0.691} & \underline{0.682} & \textbf{139.802} & 1.000 \\
\bottomrule
\end{tabular}%
}
\caption{\textbf{Full results across diverse Sage model families.}
We use \texttt{google/gemma-2-9b-it}, \texttt{mistralai/Mistral-7B-Instruct-v0.3}, and \texttt{Qwen/Qwen2.5-7B-Instruct} for Gemma, Mistral, and Qwen, respectively.
System Accuracy follows the ternary system-level metric used in the main text.
\textbf{Bold} and \underline{underlined} values denote the best and second-best results within each model-family--dataset block; for latency, lower is better. Speedup is not ranked.}
\label{tab:full_results-models}
\end{table*}

\end{document}